\documentclass{article}
\usepackage{natbib}
\usepackage[left=2cm,right=2cm, top=3cm]{geometry}
\usepackage[parfill]{parskip}
\usepackage[utf8]{inputenc}
\usepackage[T1]{fontenc}
\usepackage{amssymb}
\usepackage{amsmath}
\usepackage{amsbsy}
\usepackage{amsfonts}
\usepackage[nice]{nicefrac}
\usepackage{breqn}
\usepackage{mathtools}
\usepackage{graphicx}
\usepackage{subcaption}
\usepackage{svg}
\usepackage{tikz}
\usepackage{tikz-cd}
\usepackage{pgfplots}
\usepackage{float}
\usepackage{dblfloatfix}
\usepackage[hidelinks]{hyperref}
\usepackage{booktabs}
\usepackage{microtype}
\usepackage{multirow}
\usepackage{booktabs}
\usepackage{adjustbox}
\usepackage{soul}
\usepackage{wrapfig}
\usepackage{enumitem}
\usepackage{algorithm}
\usepackage{algorithmic}
\usepackage{siunitx}
\allowdisplaybreaks
\pgfplotsset{compat=1.16}

\title{F\textsuperscript{3}NO: Frequency-Decomposed Finite-Time Flow-map \\Neural Operators with
Cross-Scale Conditioning}
\author{
Fan Wu\\
\small Arizona State University\\
\small Tempe, AZ\\
\small\texttt{fanwu8@asu.edu}
\and
Cheng Jing\\
\small Arizona State University\\
\small Tempe, AZ\\
\small\texttt{cjing6@asu.edu}
\and
Kookjin Lee\\
\small Arizona State University\\
\small Tempe, AZ\\
\small\texttt{klee263@asu.edu}
}
\date{}

\begin{document}
\maketitle
\begin{abstract}
  \noindent Neural operators enable fast PDE forecasting, but repeated predictions accumulate errors and fine-scale structures remain difficult to resolve. We introduce a frequency-decomposed finite-time flow-map neural operator (F\textsuperscript{3}NO) that leverages updated low-frequency features to guide nonlinear refinement of high-frequency information. Within each layer, this cross-scale conditioning connects global spectral processing with local detail refinement. The model directly predicts states at specified future times and adjusts the contributions of the two branches according to the prediction interval. For longer trajectories, it combines parallel predictions within short temporal segments with recursive propagation between segments. Experiments on five PDE benchmarks demonstrate improved forecasting accuracy over autoregressive and direct-prediction baselines. Ablations show that frequency-decomposed refinement can improve accuracy with fewer parameters, while the benefits of segmentation depend on spatial resolution and dynamical regime.

\end{abstract}

\section{Introduction}


Neural operators (NOs)~\cite{lu2021learning,kovachki2023neural}, particularly Fourier neural operators (FNOs) and their variants~\cite{li2020fourier,tran2021factorized,guibas2021efficient,rahman2022u,wen2022u,raonic2023convolutional,liu2024neural,jing2026extending}, provide fast surrogate models for learning function-to-function mappings in scientific machine learning~\cite{baker2019workshop,karniadakis2021physics}. Their ability to process spatial fields makes them well suited to predicting the evolution of systems governed by partial differential equations (PDEs). Accurate prediction over long horizons, however, remains challenging for nonlinear and chaotic dynamics. A common approach is autoregressive rollout, which repeatedly applies a one-step predictor at a fixed time interval $\Delta_t$. Errors in predicted states are then propagated through subsequent predictions, potentially causing substantial long-horizon drift. Training through long rollouts can also incur high memory costs and difficult optimization due to the unrolled computation graph~\cite{brandstetter2022message}.


Several approaches address these temporal prediction challenges. Memory-augmented neural operators incorporate temporal history to improve recurrent predictions~\cite{buitrago2025benefits}. Finite-time flow-map learning instead predicts a future state directly from a source state and the elapsed time between them~\cite{hirsch2013differential,qin2019data,churchill2023flow}. Time-conditioned neural operators implement this capability by modulating intermediate features according to the requested prediction interval~\cite{herde2024poseidon}. Such operators support both direct prediction and recursive composition over shorter intervals. These options provide flexibility in how a trajectory is predicted, but accurate finite-time prediction still requires representing the spatial structures that evolve over each interval.


Frequency-dependent prediction errors present a particular challenge. PDE-Refiner demonstrates the importance of non-dominant spatial frequencies and improves their prediction through iterative refinement~\cite{lippe2023pde}. FreqMoE enhances high-frequency processing by transferring learned low-frequency weights to specialized frequency-domain experts~\cite{chen2025freqmoe}. These methods motivate treating different parts of the spectrum according to their roles in the prediction problem. Our frequency-resolved analysis further shows that substantial errors can remain in particular spatial frequency bands even when an FNO is conditioned on the prediction interval. Although standard FNO nonlinearities already allow interactions between frequencies, temporal conditioning alone does not specify how an updated global representation should guide the correction of finer-scale features. We therefore investigate an explicit feature pathway that uses the low-frequency update of the current input to condition nonlinear detail refinement.

We introduce F\textsuperscript{3}NO, a frequency-decomposed finite-time flow-map neural operator built on an FNO backbone. Its central component is a frequency-decomposed spectral refinement (FDSR) block. In each  layer, FDSR separates latent features into low- and high-frequency components and updates the low-frequency component through a global Fourier operator. A local nonlinear branch then processes the high-frequency component together with the updated low-frequency features. This conditioning acts on the features of the current input within each layer: the global spectral update directly informs the local detail correction. It differs from transferring weights between frequency experts or iteratively refining an output prediction. A lead-time-dependent fusion mechanism combines the two branches, while feature-wise affine modulation conditions the layer on the prediction interval. By processing high-frequency information through a shared local branch, FDSR also avoids allocating separate spectral weights to those modes.

The prediction interval introduces a complementary consideration. Longer intervals reduce the number of recursive transitions but can make the finite-time mapping harder to approximate; shorter intervals simplify each prediction while requiring more frequent propagation of predicted states~\cite{luchtenburg2014long}. We use segmented training and execution to examine this trade-off with F\textsuperscript{3}NO. The operator is trained to predict over bounded temporal intervals, and long trajectories are generated by propagating predicted states across segment boundaries. Within each segment, intermediate states are predicted directly from the segment's initial state and can be evaluated in parallel. FDSR thus determines how spatial information is processed, while segmentation determines the temporal interval over which that processing is learned and applied.

We evaluate F\textsuperscript{3}NO on the 1D Kuramoto--Sivashinsky equation (KSE)~\cite{kuramoto1978diffusion,ashinsky1988nonlinear}, 1D Burgers' equation (BE)~\cite{burgers1948mathematical}, 2D Navier--Stokes equation (NSE)~\cite{temam2001navier}, 2D shallow-water equation (SWE)~\cite{takamoto2022pdebench}, and 3D Rayleigh--Taylor Instability (RTI)~\cite{ohana2024well}. Our experiments assess forecasting accuracy, parameter efficiency, and the effects of spatial resolution and prediction interval.

Our contributions are summarized as follows:
\begin{itemize}[leftmargin=6mm,topsep=0pt,itemsep=0pt]
    \item We develop FDSR, an operator-layer architecture in which updated low-frequency features directly condition nonlinear refinement of high-frequency information, with lead-time-dependent fusion of the spectral and local branches.
    \item We investigate segmented training and execution for finite-time neural operators, examining the trade-off between prediction interval and recursive propagation while enabling parallel prediction within each segment.
    \item We evaluate the framework across five PDE benchmarks and use architectural and execution-mode ablations to characterize when frequency refinement and segmentation improve prediction accuracy and efficiency.
\end{itemize}
\section{Technical Background}
\subsection{Neural Operators}

NOs have emerged as a powerful data-driven paradigm for approximating function-to-function mappings, offering a resolution-invariant alternative to traditional numerical solvers~\cite{kovachki2023neural}. Formally, let $\mathcal{A}$ and $\mathcal{U}$ be separable Banach spaces of functions defined on a bounded domain $D \subset \mathbb{R}^d$. We seek to learn a non-linear operator $G^\dagger: \mathcal{A} \rightarrow \mathcal{U}$ that maps an input function $a \in \mathcal{A}$ to an output solution $u \in \mathcal{U}$. NOs approximate this by constructing a parametric surrogate $G_\theta: \mathcal{A} \to \mathcal{U}$, where parameters $\theta$ are optimized by minimizing a loss functional over a dataset of pairs $\{(a^{(i)}, u^{(i)})\}_{i=1}^N$.

\textbf{Fourier Neural Operators (FNOs).} 
A prominent instantiation of this framework is the FNOs~\cite{li2020fourier}, which parameterizes the integral kernel in the frequency domain. The architecture typically lifts the input $a(x)$ to a high-dimensional representation $v_0(x) = P(a(x))$ via a local linear transformation $P$. This representation is then evolved through a sequence of $L$ layers. Layer $\ell$ updates the features by combining a global integral operator $\mathcal{K}$ with a local linear transformation $W$:
\begin{equation}
v^{(\ell+1)}(x)
=
\sigma\Big(
W^{(\ell)}v^{(\ell)}(x)
+
\mathcal{K}^{(\ell)}(\phi)v^{(\ell)}(x)
\Big),\ \forall x\in D
\end{equation}
where $\sigma$ is a pointwise non-linear activation function.  FNOs define $\mathcal{K}$ as a convolution operator, exploiting the convolution theorem to compute the integral efficiently in the Fourier domain:
\begin{equation}
    \mathcal{K}^{(\ell)}(\phi) v^{(\ell)}(x) = \mathcal{F}^{-1} \Big( R_\phi \cdot \mathcal{F}(v^{(\ell)}) \Big)(x).
    \label{eq:fno_kernel}
\end{equation}
Here, $\mathcal{F}$ and $\mathcal{F}^{-1}$ denote the Fourier transform and its inverse, and $R_\phi$ is a learnable complex-valued weight tensor that filters the spectral modes.

\subsection{Flow-map Learning}
\label{sec:flowmap_learning}

Neural differential equation models parameterize the velocity field $f_t(u)$ of a dynamical system, $\dot{u}_t = f_t(u_t;\theta)$, and recover trajectories by numerically integrating the learned vector field~\cite{chen2018neural,kidger2020neural}. While expressive, this approach incurs substantial computational cost at inference time and inherits the numerical stability challenges associated with time integration.

Flow-map learning offers an alternative formulation that directly models the finite-time evolution of the system without explicitly representing its instantaneous dynamics~\cite{bilovs2021neural,boffi2025build}. The flow-map 
$F : [0,1]^2 \times \mathcal{D} \rightarrow \mathcal{D}$ is defined as the  operator satisfying the  condition
\begin{equation}
F_{s,t}(u_s;\theta) = u_t, \qquad \forall (s,t) \in [0,1]^2,
\label{eq:flowmap}
\end{equation}
which maps a state $u_s$ at time $s$ directly to the state $u_t$ at time $t$. By learning this operator from observational data, flow-map-based methods approximate the finite-time solution operator of the underlying dynamical system, enabling direct long-horizon prediction without recursive rollout.

\textbf{Time-conditioned feature modulation.}
Time-conditioned FNOs incorporate the prediction interval $\Delta_t$ as an additional input, allowing a single operator to predict states at different lead times~\cite{herde2024poseidon}. One implementation uses time-conditioned instance normalization (TCIN) to modulate intermediate features through feature-wise linear modulation (FiLM)~\cite{perez2018film}:
\begin{equation}\label{eq:tcin_layer}
   v^{(\ell+1)}(x)
    = \sigma \circ \mathrm{IN}_{\Delta_t}
    \Big(
        W^{(\ell)}v^{(\ell)}(x)
        + \mathcal{K}^{(\ell)}(\phi)v^{(\ell)}(x)
    \Big),
\end{equation}
where
$\mathrm{IN}_{\Delta_t}(v(x))
= \gamma(\Delta_t)\odot\mathrm{IN}(v(x))+\beta(\Delta_t)$,
and the per-channel scale $\gamma(\Delta_t)$ and shift $\beta(\Delta_t)$ are generated by a small MLP from the prediction interval.
For autonomous dynamics, the trained operator can be evaluated directly at a desired lead time or composed over shorter intervals to generate a longer trajectory~\cite{herde2024poseidon}.

\section{Methods: F\textsuperscript{3}NO}
\label{Sec: Methodology}

\begin{figure*}[t]
  \centering
  \includegraphics[scale=0.35]{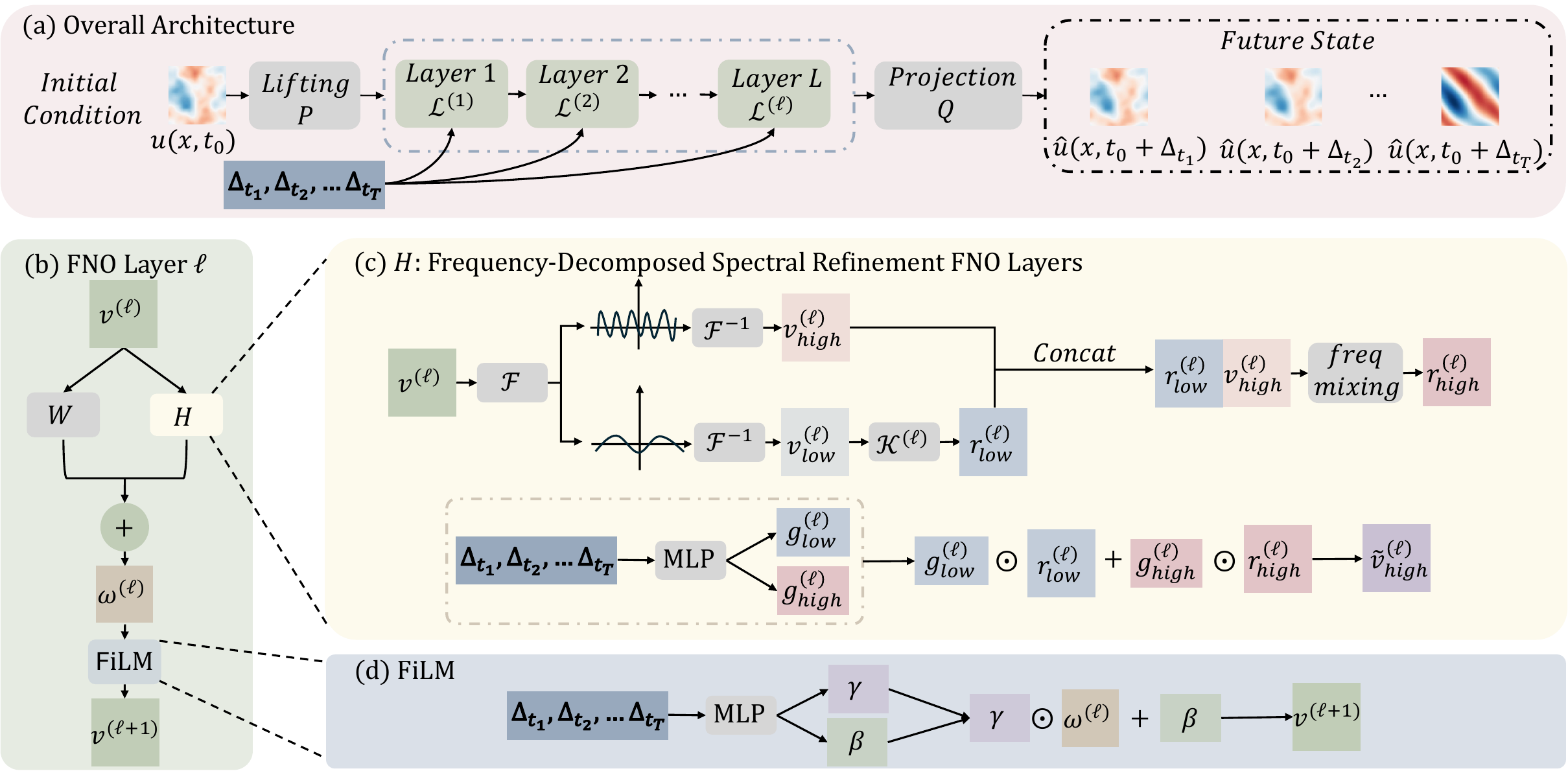}
  \caption{
\textbf{Overall architecture.}
\textbf{(a)} Given a source field $u(\mathbf{x},t_0)$ and query time offsets $\{\Delta t_i\}_{i=1}^{T}$, the model predicts future states $\hat{u}(\mathbf{x},t_0+\Delta t_i)$.
\textbf{(b)} Each layer combines a pointwise linear map with FDSR, followed by activation and FiLM conditioning.
\textbf{(c)} FDSR separates low- and high-frequency features, uses the updated low-frequency features to condition local detail refinement, and combines the branches through time-dependent gates.
\textbf{(d)} FiLM applies channel-wise scale and shift parameters conditioned on the prediction interval.}
\label{fig:model_architecture}
\vspace{-1em}
\end{figure*}

F\textsuperscript{3}NO predicts a future PDE state from a source state and a specified time interval. Its central component is frequency-decomposed spectral refinement (FDSR), which uses updated low-frequency features to condition a local nonlinear correction of high-frequency information. We first define the prediction parameterization, then describe the FDSR layers and their segmented training and execution. Figure~\ref{fig:model_architecture} summarizes the architecture.

\subsection{Finite-Time Prediction Parameterization}
\label{sec:finite_time_prediction}

For autonomous dynamics, we parameterize the learned flow-map using the source state $u_s$ and time displacement $\Delta_t=t-s$. Following residual flow-map formulations~\cite{bilovs2021neural,boffi2025build,boffi2025flow}, we write
\begin{equation}
\widehat{u}_{s+\Delta_t}
=
\widehat{F}_{\theta}(u_s,\Delta_t)
=
u_s+\frac{\Delta_t}{h_{\mathcal D}}
G_{\theta}(u_s,\Delta_t),
\label{eq:operator_forward}
\end{equation}
where $G_\theta$ is an FNO-based operator and $h_{\mathcal D}>0$ is the dataset snapshot spacing. This parameterization enforces
$\widehat{F}_{\theta}(u_s,0)=u_s$ by construction. The factor $\Delta_t/h_{\mathcal D}$ expresses the prediction interval in units of snapshot spacing. Since $G_\theta$ depends on $\Delta_t$, it represents a normalized finite-time displacement rather than an instantaneous time derivative. Training minimizes the discrepancy between the reconstructed state $\widehat{u}_{s+\Delta_t}$ and the corresponding observation.

An FNO trained for fixed-step rollout reaches later times through repeated application of the same operator. To implement the finite-time flow-map formulation above with this backbone, we condition its hidden features on the requested time displacement $\Delta_t$ using FiLM. At each FNO layer $\ell$, a learned conditioning network maps $\Delta_t$ to channel-wise scale $\gamma\left(\Delta t\right)$ and shift $\beta\left(\Delta t\right)$ coefficients, such that
\begin{equation}
    v^{(\ell+1)} = \gamma^{(\ell)}(\Delta_t)\odot\sigma\Big(W^{(\ell)}v^{(\ell)} + \mathcal{K}^{(\ell)}v^{(\ell)}\Big) +\beta^{(\ell)}(\Delta_t).
    \label{eq:flow_map_film}
\end{equation}
Unlike TCIN~\cite{herde2024poseidon}, this modulation is applied without instance normalization.

\subsection{FDSR: Frequency-Decomposed Spectral Refinement}
\label{sec:frequencydecomposed}

Temporal modulation specifies the requested prediction interval but leaves the spatial processing structure of the underlying FNO unchanged. FDSR introduces an explicit dependency between its spectral and local branches: the low-frequency features are updated first, and this update then conditions nonlinear processing of the high-frequency features. The resulting block $H^{(\ell)}(v^{(\ell)},\Delta_t)$ replaces the standard spectral term in each FNO layer.

\textbf{Frequency decomposition.}
Let $v^{(\ell)}\in\mathbb{R}^{C\times S}$ denote the latent feature field, with $C$ channels and spatial resolution $S$.\footnote{We describe the 1D case; higher-dimensional inputs use the corresponding multidimensional Fourier transform.}
Using a Fourier low-pass projector $\mathcal P_{\le m}$ with cutoff $m$, we form
$v_{\mathrm{low}}^{(\ell)}=\mathcal P_{\le m}v^{(\ell)}$
and
$v_{\mathrm{high}}^{(\ell)}=(I-\mathcal P_{\le m})v^{(\ell)}$.
The projector applies a binary mask in Fourier space and reconstructs the retained component through the inverse transform.

\textbf{Conditioned nonlinear refinement.}
Nonlinear interactions couple coarse and fine spatial scales, making high-frequency evolution dependent on the low-frequency state~\cite{foias1988modelling}. We provide this context by applying the global spectral operator to the low-frequency component and passing its updated representation to the local nonlinear branch:
\begin{equation}
\begin{aligned}
r_{\mathrm{low}}^{(\ell)}
&=
\mathcal K^{(\ell)}v_{\mathrm{low}}^{(\ell)},\\
r_{\mathrm{high}}^{(\ell)}
&=
A_2^{(\ell)}
\sigma\!\left(
A_1^{(\ell)}
\left[
v_{\mathrm{high}}^{(\ell)},
r_{\mathrm{low}}^{(\ell)}
\right]
\right).
\end{aligned}
\label{eq:fdsr_branches}
\end{equation}
Here, $[\cdot,\cdot]$ denotes channel-wise concatenation, producing a $2C$-channel input, and $A_1^{(\ell)}$ and $A_2^{(\ell)}$ are pointwise affine maps implemented with $1\times1$ convolutions. Their weights are shared across spatial locations and can be applied at different grid resolutions.

The key dependency is that the local branch (i.e., ``freq mixing'' in Figure~\ref{fig:model_architecture}) receives the \emph{updated} low-frequency features of the same input within the same layer, $r_{\text{low}}^{(\ell)}$. This provides globally processed information for local nonlinear correction and introduces a directional inductive bias for cross-scale interaction. The subscript ``high'' identifies the branch input; its nonlinear output, $r_{\text{high}}^{\ell}$, is not restricted to high frequencies. Unlike mode-specific spectral weights, the local branch's parameter count does not grow with the number of high-frequency modes, providing a potentially  parameter-efficient way to process this information.

\textbf{Lead-time conditioning.}
A small MLP
$\rho^{(\ell)}:\mathbb{R}\rightarrow\mathbb{R}^{2}$
maps $\Delta_t$ to scalar coefficients
$g_{\mathrm{low}}^{(\ell)}(\Delta_t)$ and
$g_{\mathrm{high}}^{(\ell)}(\Delta_t)$.
These coefficients combine the spectral update and local correction:
\begin{equation}
H^{(\ell)}(v^{(\ell)},\Delta_t)
=
g_{\mathrm{low}}^{(\ell)}(\Delta_t)
r_{\mathrm{low}}^{(\ell)}
+
g_{\mathrm{high}}^{(\ell)}(\Delta_t)
r_{\mathrm{high}}^{(\ell)}.
\label{eq:fdsr_fusion}
\end{equation}
The gate learns how the relative branch contributions vary with the prediction interval, without prescribing which branch should dominate at longer lead times.

Subsequently, the flow-map implementation with FiLM modulation defined in Equation~\eqref{eq:flow_map_film} becomes
\begin{equation}
v^{(\ell+1)}
=
\gamma^{(\ell)}(\Delta_t)\odot
\sigma\Big(W^{(\ell)}v^{(\ell)}+H^{(\ell)}(v^{(\ell)},\Delta_t)\Big)
+\beta^{(\ell)}(\Delta_t).
\label{eq:fno_freq_refinement}
\end{equation}
The fusion gate controls the balance between the two branches, whereas FiLM scales and shifts individual channels of the combined representation. 

\subsection{Segmented Training and Prediction}
\label{sec:hybrid}

Long prediction intervals reduce the number of recursive transitions but can increase the difficulty of approximating each finite-time mapping. Short intervals limit this difficulty while requiring more frequent propagation of predicted states. We use segmented training and prediction to study this trade-off through temporal composition~\cite{herde2024poseidon,luchtenburg2014long}.

\textbf{Temporal segmentation.}
\label{sec:training_algo}
Let $\tau\ge1$ denote the number of snapshot intervals per segment, corresponding to a duration $\tau h_{\mathcal D}$. We partition the prediction interval $[0,T]$ into
$K=\lceil T/(\tau h_{\mathcal D})\rceil$ segments with boundaries
\[
0=t_0<t_1<\cdots<t_K=T,
\qquad
t_{k+1}-t_k\le\tau h_{\mathcal D}.
\]

\textbf{Training.}
Our default setting, Hybrid-TF, uses the ground-truth state $u_{t_k}$ as the input at each segment boundary. The model directly predicts the observed states within that segment using their respective time displacements. The loss averages snapshot-wise relative $L^2$ errors over the predicted states. This trains the operator on bounded time intervals without propagating prediction errors between training segments. We also consider Hybrid-RO, which  feeds the predicted terminal state into the next segment during training. Algorithm~\ref{alg:hybrid} in Appendix~\ref{app:implementation} details both variants. 

\textbf{Prediction.}
At inference, segment boundaries are initialized from predictions, starting with $\widehat{u}_0=u_0$:
\begin{equation}
\widehat{u}_{t_k+\Delta_t}
=
\widehat{F}_{\theta}
(\widehat{u}_{t_k},\Delta_t),
\qquad
0<\Delta_t\le t_{k+1}-t_k.
\label{eq:segment_flowmap}
\end{equation}
All requested states within a segment share the same input $\widehat{u}_{t_k}$ and can be evaluated in parallel. The terminal prediction $\widehat{u}_{t_{k+1}}$ initializes the next segment.

With $\tau\!\!=\!\!1$, this procedure reduces to prediction at each  interval; a segment spanning the entire horizon gives direct prediction from the initial state. Intermediate choices restrict recursive state propagation to segment boundaries. Their accuracy depends on both the approximation error over each interval and the amplification of boundary errors, which we examine experimentally across segment lengths and dynamical regimes.

\section{Experiments}
\label{sec:experiments}

We now empirically demonstrate how the proposed components in F\textsuperscript{3}NO resolve the issue of prediction error growth across time and frequency scales, leading to general improvements to predictive accuracy across a wide range of PDE systems.

\subsection{Experimental Setup}
\label{sec:exp_setup}

\textbf{Benchmark.} We evaluate F\textsuperscript{3}NO on the  
1D Burgers' equation (BE), 1D Kuramoto--Sivashinsky equation (KSE),  2D Navier--Stokes equation (NSE), 2D Shallow--Water equation (SWE), and 3D Rayleigh--Taylor Instability (RTI). These datasets cover nonlinear convection, dissipative dynamics, and chaotic time evolution. Data generation and prediction settings are detailed in Appendix~\ref{sec:appendix_benchmark}. 

\textbf{Baselines.} As baselines of comparison, we consider methods from two broad classes: rollout and flow-map. Rollout-based methods are trained either in pure rollout mode or using teacher forcing (TF). 
We evaluate FNO~\cite{li2020fourier}, FFNO~\cite{tran2021factorized}, UNet~\cite{gupta2022towards}, 
S4FFNO~\cite{buitrago2025benefits}, PDE-Refiner~\cite{lippe2023pde}, and FreqMoE~\cite{chen2025freqmoe}. 
Flow-map-based baselines directly learn a finite-time solution operator that maps between states separated by arbitrary time gaps, including a simple MLP, DeepONets (DONs)~\cite{lu2021learning}, TCIN~\cite{herde2024poseidon}, and HyperFNO~\cite{alesiani2022hyperfno}
.

\textbf{Training and testing.} Given a discrete-time trajectory $\{u_{t}\}_{t=0}^{T}$, the goal of neural operator training is to predict the full trajectory from the initial state $u_0$. 
For all methods, the models make predictions on a trajectory $\{\hat u_{t}\}_{t=1}^{T}$. For training, we compute a mean solution-snapshot-wise  normalized error. For testing, we provide a novel initial condition to the trained model and collect a predicted trajectory, $\hat{u}_{1:T}$. Then we evaluate the model performance using the normalized root mean square error (nRMSE), $\mathrm{nRMSE}
= \frac{1}{n_{\text{test}}}\sum_{k=1}^{n_{\text{test}}}
{\left\|u_{1:T}^{(k)}-\hat{u}_{1:T}^{(k)}\right\|_{2}} / 
{\left\|u_{1:T}^{(k)}\right\|_{2}}$, where ${u}_{1:T}$ refers to the ground-truth trajectory, the superscript $^{(k)}$ indicates the $k$-th trajectory, and $n_{\text{test}}$ refers to the number of test trajectories. 

Unless otherwise stated, F\textsuperscript{3}NO denotes the FNO backbone equipped with FDSR and FiLM conditioning, trained with Hybrid-TF and evaluated using segmented flow-map execution
. The segment length is fixed to $\tau=4$ for KSE, NSE, and RTI, $\tau=10$ for SWE, and $\tau=19$ for BE. The low-frequency spectral branch uses approximately three-quarters of the available Fourier modes on BE and KSE, 16 modes on NSE and SWE, and 12 modes on RTI.


\subsection{In-Depth Analysis of KSE Forecasting}

We use KSE as a case study to examine prediction errors across time and frequency, inference efficiency, and the contribution of cross-scale conditioning, with additional ablations on NSE.
\begin{figure}[t]
\vspace{-1.2em}
    \centering
    \includegraphics[width=0.95\linewidth]{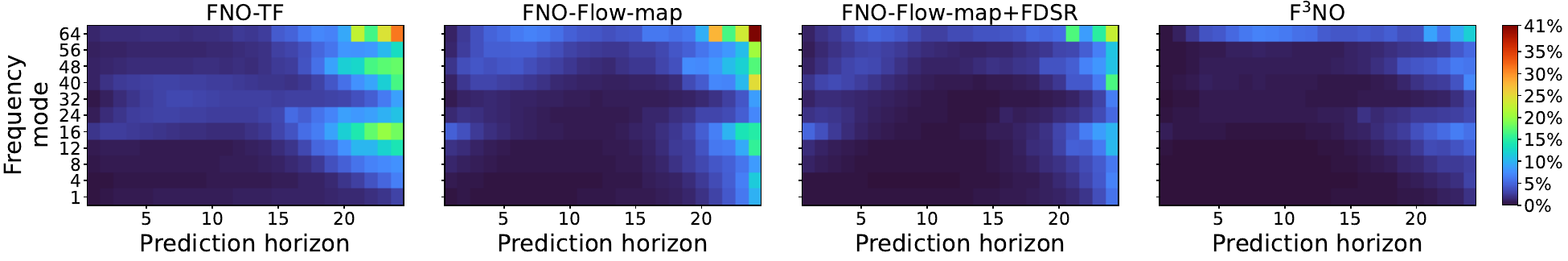}
    \vspace{-0.6em}
    \caption{[KSE, $\nu=0.075$, $S=128$] Frequency-resolved prediction error as a function of prediction horizon. Each cell reports the time-instantaneous (non-cumulative) per-mode relative $L^2$ error (\%). 
    }
    \label{fig:freqloss}
\end{figure}
\textbf{Prediction errors across time and frequency.}
Figure~\ref{fig:freqloss} compares four variants at $\nu=0.075$ and $S=128$: FNO-TF is trained with teacher forcing and evaluated autoregressively. FNO-Flow-map, (based on the time-conditioning described in eq.~\eqref{eq:flow_map_film} in Sec.~\ref{sec:finite_time_prediction}), directly predicts future states from the initial state at different lead times. FNO-Flow-map+FDSR augments this direct predictor with frequency-decomposed spectral refinement (FDSR). F\textsuperscript{3}NO further builds on this model by adopting segmented execution.

This comparison provides a motivating example for our approach. FNO-TF errors grow with the forecast horizon, consistent with recursive error accumulation. FNO-Flow-map reduces errors over part of the horizon, yet substantial errors remain at longer lead times. These observations motivate improving the finite-time predictor while controlling how far each prediction advances, balancing the difficulty of long-range mappings against repeated propagation of predicted states.

Figure~\ref{fig:freqloss} also shows that these prediction errors are not distributed uniformly across frequencies. In both FNO-TF and FNO-Flow-map, the errors vary noticeably across Fourier modes, with clear error bands emerging as the horizon increases. Thus, beyond temporal error growth, there is also a frequency-dependent prediction difficulty. When FDSR is added, these cross-frequency discrepancies become more evenly controlled across Fourier modes, and the dominant error bands are weakened. 
This indicates that explicitly separating and coupling different frequency components improves the quality of the finite-time predictor.

Although FDSR improves prediction accuracy, FNO-Flow-map+FDSR still exhibits substantial errors at longer lead times in this example. This motivates limiting the temporal span of each prediction through segmented training and execution. Figure~\ref{fig:freqloss} shows that F\textsuperscript{3}NO further reduces these late-horizon errors, illustrating the potential benefit of combining spectral refinement with shorter prediction intervals.

\begin{wrapfigure}{r}{0.48\linewidth}
\vspace{-1.6em}
    \centering
    \includegraphics[width=\linewidth]{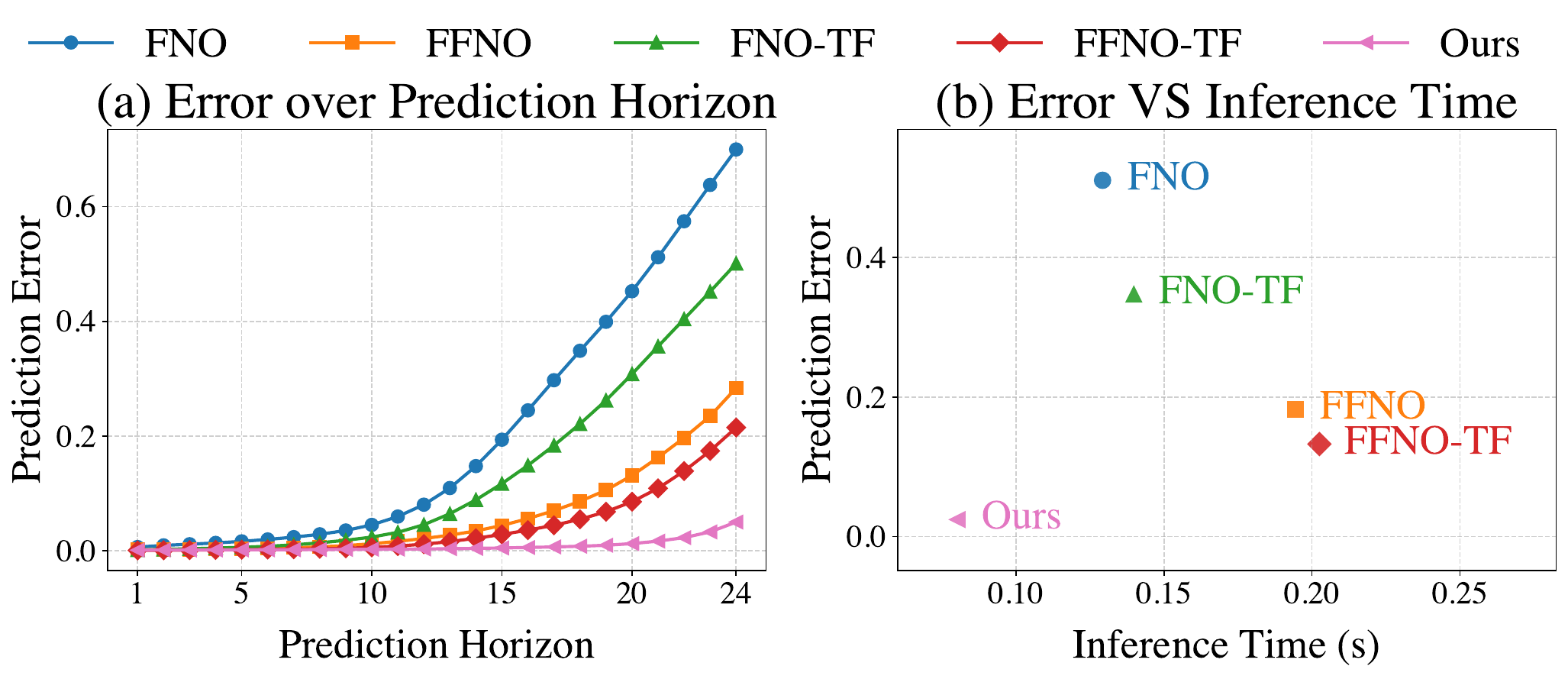}
    \vspace{-1.2em}
    \caption{[KSE, $\nu=0.075$, $S=64$] Prediction error measured as a function of prediction horizon. The prediction error is reported as the time-instantaneous (non-cumulative) relative $L^2$ error at each time step.}
    \label{fig:timeVSloss}
\vspace{-1.2em}
\end{wrapfigure}
\textbf{Forecasting accuracy and inference efficiency.}
We next assess overall forecasting performance at the coarser resolution $S=64$ on KSE with $\nu=0.075$. Figure~\ref{fig:timeVSloss}a shows prediction error as a function of the forecast horizon. Rollout-based methods exhibit rapid error growth and become unstable as the horizon increases, reflecting compounding errors from autoregressive prediction. In contrast, the F\textsuperscript{3}NO maintains controlled error growth and preserves predictive accuracy over substantially longer horizons, indicating improved long-horizon stability.

\begin{table*}[t]
  \caption{Relative $L^2$ error (\%) with mean $\pm$ standard deviation for varying datasets and varying PDE parameters, with a resolution $S=128$ for BE/KSE, a resolution $S=64\times64$ for NSE and SWE, and a resolution $S=32\times32\times32$ for RTI. Best and second-best performance are bolded and underlined, respectively. 
  }
  \label{tab:varying_data_S128}
  \vspace{-0.6em}
  \centering
      \scriptsize
    \setlength{\tabcolsep}{4pt}
  \begin{tabular}{lccccccccc}
    \toprule
     & \multicolumn{1}{c}{BE} & \multicolumn{3}{c}{KSE} & \multicolumn{3}{c}{NSE}  & \multicolumn{1}{c}{SWE} &  \multicolumn{1}{c}{RTI}       \\
         & $\nu=0.001$ & $\nu=0.075$ & $\nu=0.1$ & $\nu=0.125$ & $\nu=10^{-5}$ & $\nu=10^{-4}$ & $\nu=10^{-3}$ & $r\sim\mathcal{U}(0.3,0.7)$ & $A_t=0.25$\\
    \midrule
    \multicolumn{8}{l}{Rollout} \\
    \cmidrule(lr){2-10}
    FNO & 
    7.64$\pm$1.09 & 33.23$\pm$0.76 & 16.21$\pm$1.89 & 12.45$\pm$1.93 & 2.27$\pm$0.12 & 2.61$\pm$0.20 & 1.42$\pm$0.32 & 3.67$\pm$0.26 & 6.00$\pm$0.18 \\
    FFNO  & 
    4.02$\pm$0.43 & 8.02$\pm$0.85 & 3.42$\pm$0.28 & 2.10$\pm$0.55 & \underline{\smash{1.53$\pm$0.03}} & \underline{\smash{0.76$\pm$0.06}} & 0.52$\pm$0.02 & 2.52$\pm$0.39 & 5.73$\pm$0.31  \\
    FNO-TF & 
    4.72$\pm$0.01 & 3.64$\pm$0.08 & 1.48$\pm$0.10 & 0.80$\pm$0.04 & 1.96$\pm$0.03 & 1.02$\pm$0.03 & 0.67$\pm$0.03 & 2.02$\pm$0.11 & 5.93$\pm$0.11\\
    FFNO-TF & 
    4.26$\pm$0.24 & \underline{\smash{1.26$\pm$0.06}} & 0.47$\pm$0.04 & 0.35$\pm$0.01 & 1.60$\pm$0.01 & 1.23$\pm$0.06 & 0.45$\pm$0.02 & 0.56$\pm$0.11 & 5.34$\pm$0.13\\
    S4FFNO & 
    \underline{\smash{1.85$\pm$0.05}} & 3.17$\pm$0.18 & 2.24$\pm$0.29 & 0.47$\pm$0.04 & 3.37$\pm$2.41 & 4.71$\pm$2.22 & 0.70$\pm$0.10 & 0.67$\pm$0.17 & 5.17$\pm$0.03\\
    UNET-TF & 
    5.45$\pm$0.70 & 9.04$\pm$0.98 & 4.07$\pm$0.38 & 2.50$\pm$0.06 & 5.17$\pm$0.78 & 3.92$\pm$0.33 & 1.22$\pm$0.12 & 27.78$\pm$8.94 & 7.23$\pm$0.38\\
    PDE-Refiner & 7.25$\pm$0.71 & 1.61$\pm$0.25 & \underline{\smash{0.44$\pm$0.12}} & \underline{\smash{0.25$\pm$0.01}} & 1.58$\pm$0.01 & 0.91$\pm$0.02 & \underline{\smash{0.31$\pm$0.01}} & 0.28$\pm$0.01 & 5.60$\pm$0.16\\
    FreqMoE &5.78$\pm$0.24 &2.95$\pm$0.20&1.15$\pm$0.05&0.74$\pm$0.01&1.96$\pm$0.03&1.10$\pm$0.04&0.66$\pm$0.01 & 2.53$\pm$0.47 &5.53$\pm$0.29\\
    \midrule
    \multicolumn{8}{l}{Flow-map} \\
    \cmidrule(lr){2-10}
    MLP & 
    50.73$\pm$0.04 & 88.18$\pm$0.01 & 75.27$\pm$0.08 & 67.59$\pm$0.01 & 29.17$\pm$0.02 & 30.26$\pm$0.06 &26.72$\pm$0.03 & 4.77$\pm$0.12 & 5.36$\pm$0.04
    \\
    DON & 
    24.11$\pm$0.49 & 88.24$\pm$0.04 & 75.02$\pm$0.16 & 65.95$\pm$0.14 & 27.57$\pm$0.31 & 24.86$\pm$0.16 &21.33$\pm$0.11 & 9.23$\pm$0.04 & 5.42$\pm$0.02\\
    hyperFNO & 
    2.23$\pm$0.08 & 4.89$\pm$0.33 & 1.45$\pm$0.13 & 0.71$\pm$0.03 & 9.74$\pm$0.34 & 7.64$\pm$0.11 & 1.20$\pm$0.13 & \underline{\smash{0.13$\pm$0.01}} & 5.14$\pm$0.01\\
    FNO-TCIN& 
    91.68$\pm$0.62 & 8.24$\pm$0.05 & 3.38$\pm$0.03 & 2.00$\pm$0.01 & 9.72$\pm$0.52 &7.71$\pm$0.17 & 2.19$\pm$0.09 & 0.25$\pm$0.05 & 5.15$\pm$0.03\\
    FFNO-TCIN& 
    1.88$\pm$0.07 & 2.84$\pm$0.01 & 1.01$\pm$0.04 & 0.59$\pm$0.09 & 3.03$\pm$0.14 & 3.25$\pm$0.07 & 1.73$\pm$0.13 & 0.26$\pm$0.00 & \underline{\smash{5.12$\pm$0.02}}\\
    \midrule
    F\textsuperscript{3}NO (ours) & 
    \textbf{1.37$\pm$0.11} & 
    \textbf{1.15$\pm$0.06} & 
    \textbf{0.32$\pm$0.00} & 
    \textbf{0.17$\pm$0.01} & 
    \textbf{1.44$\pm$0.04} & 
    \textbf{0.58$\pm$0.02} & 
    \textbf{0.22$\pm$0.00} &
    \textbf{0.07$\pm$0.01} & 
    \textbf{4.83$\pm$0.08}
    \\
    \bottomrule
  \end{tabular}
  \vspace{-1.5em}
\end{table*}

Figure~\ref{fig:timeVSloss}b illustrates the trade-off between inference time and prediction error. 
F\textsuperscript{3}NO achieves lower error while also being faster than rollout-based baselines. 
This efficiency gain is partly due to the execution paradigm: autoregressive rollouts require sequential reuse of predicted states as inputs at each time step, preventing parallelization. 
In contrast, the flow-map formulation supports batched inference over multiple time gaps, i.e., evaluating $G_{\theta}(u_s, \Delta_t)$ for a batch of $\Delta_t$ values in a single forward pass. 
This enables efficient parallel computation and results in substantially reduced inference time without sacrificing accuracy.

\textbf{Ablation on cross-scale conditioning.} To isolate the contribution of cross-scale conditioning, we compare FDSR with high-frequency-only refinement, conditioning on the original low-frequency features, and a pointwise MLP with comparable parameter count applied to undecomposed features. Keeping the low-frequency spectral branch, temporal conditioning, and segmented training and prediction settings fixed, FDSR achieves the lowest error among these variants across the tested KSE viscosities. Additional ablations on NSE show the same ranking, supporting the benefit of updated low-frequency conditioning beyond additional local nonlinear capacity. Detailed comparisons are provided in Appendix~\ref{app:ablation-FDSR}, Tables~\ref{tab:kse_lowfreq_ablation128} and~\ref{tab:nse_lowfreq_ablation128}.

\subsection{Performance across PDE benchmarks.} 
Following the preceding analysis of temporal and spectral prediction errors, we evaluate whether these result in more accurate predictions across benchmark PDEs.
Table~\ref{tab:varying_data_S128} summarizes the performance of the proposed method, F\textsuperscript{3}NO, alongside all considered baselines. 
Across all benchmarks in Table~\ref{tab:varying_data_S128}, F\textsuperscript{3}NO achieves the lowest error, corresponding to approximately 
6\%-46\%
relative error reduction over baselines.

\textbf{Varying viscosity, $\nu$.}
In KSE and NSE, decreasing the viscosity \(\nu\) brings more chaos in the dynamics and we assess whether the proposed method remains robust as the underlying dynamics become more chaotic.
As shown in Table~\ref{tab:varying_data_S128}, all methods exhibit increasing error as \(\nu\) decreases; however, the magnitude and rate of degradation vary substantially across approaches. 
As the viscosity decreases and the dynamics become increasingly chaotic, both rollout- and flow-map-based baselines generally exhibit noticeable performance degradation across all benchmark problems. 
Across all viscosity settings, the proposed method achieves approximately 
6\%-46\%
relative improvement over baselines. This behavior highlights the key advantage of segmented execution: by restricting each flow-map application to a bounded temporal interval while retaining global efficiency within segments, the method effectively balances stability and expressivity in regimes where either rollouts or single-step flow-maps prediction alone become unreliable.


\begin{wraptable}{r}{0.45\linewidth}
\vspace{-1.5em}
  \caption{Relative $L^2$ error (\%) with mean $\pm$ standard deviation for varying spatial resolutions $S\in\{32,64,128\}$ with the fixed $\nu=0.075$.}
  \label{tab:res075}
  \vspace{-0.8em}
  \centering
  \scriptsize
  \setlength{\tabcolsep}{4pt}
    \begin{tabular}{lrrr}
      \toprule
          & \multicolumn{1}{c}{32} & \multicolumn{1}{c}{64} & \multicolumn{1}{c}{128} \\
      \midrule
      \multicolumn{4}{l}{Rollout} \\
      \cmidrule(lr){2-4}
      FNO      & 78.96$\pm$6.06 & 50.76$\pm$2.91 & 33.23$\pm$0.76 \\
      FFNO     & 47.89$\pm$1.23 & 18.31$\pm$1.32 & 8.02$\pm$0.85 \\
      FNO-TF   & 101.68$\pm$3.85 & 37.24$\pm$0.79 & 3.64$\pm$0.08 \\
      FFNO-TF  & 72.93$\pm$1.14 & 13.20$\pm$0.20 & \underline{\smash{1.26$\pm$0.06}} \\
      UNET-TF  & 87.02$\pm$3.41 & 31.13$\pm$1.48 & 9.04$\pm$0.98 \\
      S4FFNO   & 63.31$\pm$8.79 & 9.90$\pm$0.21 & 3.17$\pm$0.18 \\
      PDE-Refiner & 72.72$\pm$1.48 &16.18$\pm$0.64 & 1.61$\pm$0.25\\
      FreqMoE & 49.67$\pm$1.64& 18.36$\pm$2.16&2.95$\pm$0.20\\
      \midrule
      \multicolumn{4}{l}{Flow-map} \\
    \cmidrule(lr){2-4}
      MLP               & 87.87$\pm$0.00 & 87.92$\pm$0.01 & 88.18$\pm$0.01 \\
      DON               & 99.02$\pm$3.99 & 87.89$\pm$0.27 & 88.24$\pm$0.04 \\
      hyperFNO          & \underline{\smash{6.24$\pm$0.20}} & \underline{\smash{5.05$\pm$0.21}} & 4.89$\pm$0.33 \\
      \midrule
      F\textsuperscript{3}NO (ours)
                         & \textbf{3.91$\pm$0.06}
                         & \textbf{2.95$\pm$0.10}
                         & \textbf{1.15$\pm$0.06} \\               
      \bottomrule
    \end{tabular}
\vspace{-3em}
\end{wraptable}

\textbf{Varying spatial resolution, $S$.}
Next, we conduct a varying-spatial-resolution experiment to evaluate the robustness of the proposed method under changes in the fidelity of the spatial representation. Increasing the spatial resolution introduces higher-frequency modes (i.e., spatial details), which makes learning a flow-map more challenging: the operator must accurately capture interactions across a broader spectral range while maintaining temporal stability. Furthermore, higher resolution typically benefits rollouts by reducing discretization error at each step. By fixing $\nu=0.075$ and varying the resolution in $\{32,64,128\}$, this experiment isolates the effect of spatial complexity on different temporal modeling strategies.

As shown in Table~\ref{tab:res075}, prediction error decreases for all methods as spatial resolution increases, reflecting improved representation of local dynamics.
Despite the increased spectral complexity at higher resolutions, the proposed segmented flow-map consistently achieves the lowest error across all settings, resulting in 9\%--42\% improvements compared to the second best performing models. 
While rollout-based models narrow the performance gap as resolution increases, they remain less accurate than the proposed method even at the highest resolution.
This suggests that the advantage of the proposed approach does not arise solely from coarse-grid effects, but from its ability to model finite-time evolution in a manner that remains robust as spatial complexity increases.

\begin{figure}[b]
    \centering
    \vspace{-1.5em}
    \includegraphics[width=\linewidth]{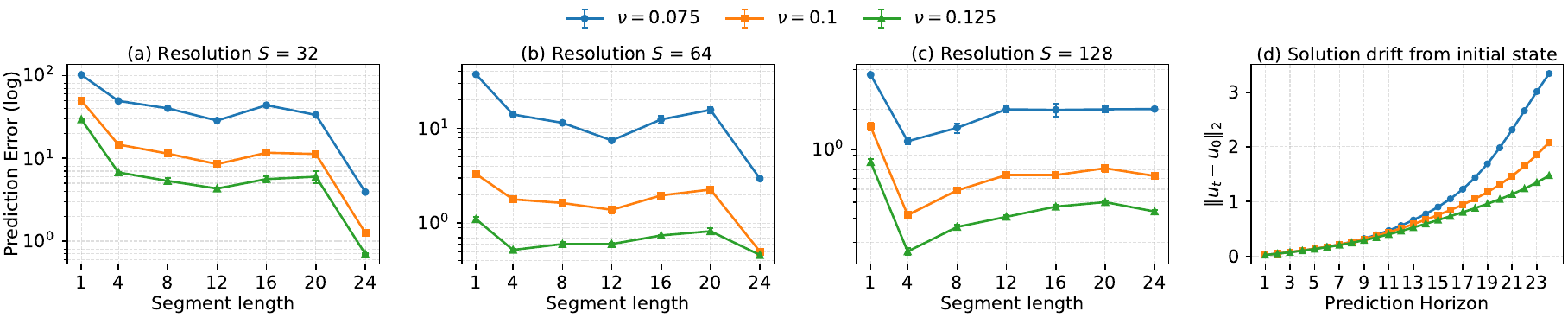}
    \vspace{-1.5em}
    \caption{(a,b,c) prediction error for the proposed model as a function of segmentation length over three spatial resolutions. (d) $\|u_t - u_0\|$ quantifies how rapidly the system evolves away from its initial condition.}
    \label{fig:prediction_errorVSseg_length}
\end{figure}

\textbf{Effectiveness of segmented flow-map for varying $\tau$.} 
We further investigate how the segmentation length $\tau$ affects performance across different dynamical regimes by varying $\tau \in \{1, 4,8,12,16,20,24\}$ for multiple viscosities $\nu$ and spatial resolutions. 
Figure~\ref{fig:prediction_errorVSseg_length} is a sensitivity analysis over the segment length $\tau$. 

Across settings, we observe a clear interaction between the degree of chaos, spatial resolution, and the optimal segmentation length. 
For higher spatial resolutions and smaller $\nu$, corresponding to more chaotic dynamics, smaller segmentation lengths tend to yield better performance, indicating that more frequent switching between rollout and flow-map execution improves stability. 
In contrast, for lower resolutions or less chaotic regimes, larger segmentation lengths are preferred, and in some cases a single global flow-map ($\tau=T$) achieves the lowest error. These observations are consistent with the trade-off underlying segmented prediction. Shorter segments make each finite-time mapping easier to learn but introduce more recursive boundary transitions, whereas longer segments reduce recursive composition at the cost of harder long-range predictions. Based on the empirical sweeps, the main results in Table~\ref{tab:varying_data_S128} use $\tau$ preferred by the empirical sweeps where $\tau=19$ for BE, $\tau=4$ for KSE, NSE, and RTI, and $\tau=10$ for SWE. 
Appendix~\ref{sec:error_propagation} provides a complementary error-propagation analysis and shows that the preferred execution intervals identified from the direct prediction error per snapshot interval agree with these empirical choices across all five benchmarks.

To further contextualize this behavior, Figure~\ref{fig:prediction_errorVSseg_length}d reports the state displacement $\|u_t - u_0\|$ as a function of time, which quantifies how rapidly the system evolves away from its initial condition. 
More chaotic regimes exhibit larger state displacements over the same horizon, indicating increased difficulty for direct long-range flow-map learning and motivating more frequent segmentation. 
Conversely, in less chaotic regimes where state evolution is milder, direct flow-map execution remains effective, explaining why larger segmentation lengths or pure flow-map execution suffice.

In Appendix~\ref{app:freq} and ~\ref{app:seg}, we present additional empirical analyses and experimental results including ablations on operation modes (rollout, flow-map, hybrid), 
hyper-parameter choices (e.g., mode cutoff parameter), etc. We also report the results of experiments performed on the irregularly sampled trajectories.

\section{Related Work}
Neural operators provide a framework for learning mappings between functions across discretizations. The FNO~\cite{li2020fourier} parameterizes integral kernels in the frequency domain. Subsequent work improves efficiency and expressivity: Factorized FNO (FFNO)~\cite{tran2021factorized} uses separable spectral layers, S4FFNO~\cite{buitrago2025benefits} incorporates temporal memory through state-space modeling, and AFNO~\cite{guibas2021efficient} introduces efficient Fourier-domain channel mixing and soft-thresholding. U-NO~\cite{rahman2022u}, U-FNO~\cite{wen2022u}, and U-Net variants~\cite{gupta2022towards} use hierarchical or complementary local processing to capture multiscale structure. HyperFNO~\cite{alesiani2022hyperfno} instead uses hypernetworks to generate operator weights conditioned on PDE parameters.

Spectral bias~\cite{rahaman2019spectral} and Fourier-mode truncation can hinder fine-scale prediction, motivating frequency-aware designs. PDE-Refiner~\cite{lippe2023pde} iteratively refines predicted states to recover non-dominant frequency information. Spectral-Refiner~\cite{cao2025spectral} fine-tunes an untruncated spatiotemporal spectral layer using a PDE-residual-based objective, while FreqMoE~\cite{chen2025freqmoe} transfers learned low-frequency weights to high-frequency experts. Our FDSR block instead decomposes latent features and uses the updated low-frequency representation of the current input to condition nonlinear refinement of the high-frequency branch within each operator layer. This distinguishes its feature-level conditioning from output-state refinement, spectral-layer fine-tuning, and weight transfer.

For time-dependent systems, neural operators are often deployed autoregressively, repeatedly mapping $u_t$ to $u_{t+\Delta_t}$ at a fixed time step. This sequential dependence can accumulate prediction errors and increase long-horizon inference latency. Neural ODEs~\cite{chen2018neural} represent continuous-time dynamics through learned derivatives but require numerical integration. Neural flows~\cite{bilovs2021neural,boffi2025flow} and time-conditioned operators~\cite{herde2024poseidon} directly parameterize finite-time mappings. Building on this formulation, we combine FDSR with segmented execution to balance finite-time mapping difficulty against recursive error propagation.

\section{Conclusion}
We introduced F\textsuperscript{3}NO, a frequency-decomposed finite-time flow-map neural operator for accurate and efficient PDE forecasting. The Frequency-Decomposed Spectral Refinement uses updated low-frequency features to guide nonlinear high-frequency refinement, explicitly connecting global spectral processing with local detail correction. This improves the representation of fine-scale dynamics while providing a parameter-efficient mechanism for processing high-frequency information. Lead-time-dependent fusion further adjusts the contributions of the low- and high-frequency branches to the prediction interval. Complementing this architecture, segmented training and prediction balance the difficulty of learning long-range mappings against recursive error accumulation. By predicting states directly within each segment and propagating predictions only across segment boundaries, the framework reduces sequential dependence and can improve long-horizon accuracy. Across five PDE benchmarks, F\textsuperscript{3}NO achieves the lowest mean prediction error across all evaluated settings. These results demonstrate the value of combining explicit cross-scale conditioning with flexible temporal composition for long-horizon prediction of complex PDE dynamics.




\bibliographystyle{plain}
\bibliography{ref}
\clearpage
\appendix

\section{Benchmark Problems}
\label{sec:appendix_benchmark}
We describe details of the benchmark PDEs considered in this study: 1D Burgers' equation, 1D Kuramoto--Sivashinsky equation, 2D  Navier--Stokes equation, 2D Shallow--Water equation, and 3D Rayleigh Taylor Instability. Table~\ref{tab:dataset_overview} summarizes the prediction settings. We use $u_t$ to denote the predicted state field at time $t$, corresponding to the solution field for BE and KSE, vorticity for NSE, water depth for SWE, and density for RTI. 

\begin{table*}[h]
\centering
      \scriptsize
    \setlength{\tabcolsep}{6pt}
\caption{Overview of the datasets and prediction settings. Autoregressive baselines learn a one-step transition, while our model learns a time-conditioned flow-map mapping.}
\label{tab:dataset_overview}

\begin{tabular}{lccccccccc}
\toprule
\multirow{2}{*}{Dataset} & \multirow{2}{*}{PDE Parameter} & \multirow{2}{*}{S} & \multirow{2}{*}{$N_t$} & \multicolumn{2}{c}{Rollout Baseline} & \multicolumn{2}{c}{Our Model} & \multirow{2}{*}{N\_{train}} & \multirow{2}{*}{N\_{test}} \\
\cmidrule(lr){5-6} \cmidrule(lr){7-8}
 &  & & &Input & Output & Input & Output \\
\midrule
1D BE & $\nu=0.001$ & 128, 64, 32 & 20 & $u_t$ & $u_{t+1}$ & $(u_t,\Delta t)$ & $u_{t+\Delta t}$ & 2048 & 1000\\
1D KSE      & $\nu\in\{0.075, 0.1, 0.125\}$ &128, 64, 32 & 25 & $u_t$ & $u_{t+1}$ & $(u_t,\Delta t)$ & $u_{t+\Delta t}$ & 2048 & 256\\
2D NSE  & $\nu\in\{10^{-5},10^{-4},10^{-3}\}$ & $64\times64$ & 20, 30, 50 & $u_t$ & $u_{t+1}$ & $(u_t,\Delta t)$ & $u_{t+\Delta t}$ & 1000 & 200\\
2D SWE & $r\sim\mathcal{U}(0.3,0.7)$ & $64\times64$ & 101 & $u_t$ & $u_{t+1}$ & $(u_t,\Delta t)$ & $u_{t+\Delta t}$ & 900 & 100\\
3D RTI & $A_t=0.25$ & $32\times32\times32$ & 25 & $u_t$ & $u_{t+1}$ & $(u_t,\Delta t)$ & $u_{t+\Delta t}$ & 9 & 2\\
\bottomrule
\end{tabular}
\end{table*}

\paragraph{1D Burgers' Equation.}
The viscous Burgers' equation (BE), parameterized by a viscosity $\nu\in\mathbb{R}_+$, is a widely used nonlinear advection-diffusion model in settings such as fluid transport, traffic-flow modeling, and other dynamic phenomena. 
The viscous BE is given by:
\begin{equation}
u_t + u\,u_x = \nu u_{xx},
\qquad (t,x)\in[0,T]\times[0,L], 
\label{eq:burgers}
\end{equation}
We use the publicly available BE benchmark from the PDEBench repository with viscosity $\nu=0.001$, provided at a spatial resolution of 1024. We perform experiments at resolution 128, 64, and 32, with final time $T\!=\!20$.

\paragraph{1D Kuramoto--Sivashinsky Equation.}

The Kuramoto--Sivashinsky Equation (KSE) is a nonlinear, dissipative partial differential equation that exhibits spatiotemporal chaos, making it a widely adopted testbed for studying long-term prediction, stability, and error accumulation in learned dynamical models. Its well-understood physical structure and extensive use in prior work allow for direct and meaningful comparison with existing NO approaches.

The KSE equation is governed by
\begin{equation}
\begin{split}
u_t + u\,u_x + u_{xx} + \nu u_{xxxx} = 0, 
\qquad u(0,x) = u_0(x),
\end{split}
\label{eq:ks}
\end{equation}
where $(t,x)\in[0,T]\times[0,L]$ with periodic boundary conditions and viscosity parameter $\nu \in \mathbb{R}_+$. The nonlinear advection term $u\,u_x$, combined with fourth-order dissipation, induces strong sensitivity to initial conditions and rapid divergence of nearby trajectories for sufficiently small $\nu$. These properties render the KSE particularly challenging for autoregressive rollouts, where local approximation errors compound exponentially over time.

We follow the data-generation protocol used in prior work~\cite{buitrago2025benefits}, simulating trajectories on a periodic domain of length $L=64$ with a reference spatial resolution of $512$. Datasets are generated for viscosity values $\nu \in \{0.075, 0.1, 0.125\}$, spanning increasingly chaotic regimes. For evaluation, the solutions are downsampled to spatial resolutions of $\{32,64,128\}$. All simulations are carried out up to a final time of 2.5 time units with a timestep of 0.1.

\paragraph{2D Navier--Stokes Equations.}
The Navier-Stokes equation (NSE) describes the motion of viscous fluids. We consider the two-dimensional Navier-Stokes equation on the unit torus in vorticity form

\begin{equation}
\begin{aligned}
\partial_t w(x,t)+u(x,t)\!\cdot\!\nabla w(x,t) &= \nu\Delta w(x,t)+f(x), && x\in(0,1)^2,\;t\in(0,T],\\
\nabla\!\cdot\!u(x,t) &= 0, && x\in(0,1)^2,\;t\in[0,T],\\
w(x,0) &= w_0(x), && x\in(0,1)^2
\end{aligned}
\label{eq:ns}
\end{equation}
Initial vorticity fields are drawn from the stationary Gaussian measure $\omega_0 \sim \mathcal{N}\!\bigl(0,\,7^{3/2}(-\Delta+49I)^{-2.5}\bigr)$, while the time‑independent forcing is fixed to $f(\mathbf{x}) = 0.1\!\bigl[\sin 2\pi(x_1+x_2)+\cos 2\pi(x_1+x_2)\bigr]$. 

We follow the data generation process of \cite{li2020fourier}. A Crank-Nicolson scheme is used with a time-step of $10^{-4}$ during data-generated process and the solution is recorded every $t=0.25$ seconds. All data are generated on a $256 \times 256$ grid and downsampled to $64 \times 64$ for both training and testing. We considered the Navier--Stokes equations for viscosities $\nu\!=\!10^{-3}$, $ 10^{-4}$, and $ 10^{-5}$ with a final time of 5 seconds, 7.5 seconds, and 12.5 seconds, respectively.

\paragraph{2D Shallow Water Equations.}
The shallow-water equations (SWE) describe free-surface flows in which the horizontal length scale is much larger than the fluid depth. We consider the two-dimensional SWE in conservative form
\begin{equation}
\begin{aligned}
    \partial_t h + \partial_x(hu) + \partial_y(hv) &= 0, \\
    \partial_t(hu)
    + \partial_x\left(u^2h + \frac{1}{2}g_rh^2\right)
    + \partial_y(uvh)
    &= -g_rh\,\partial_x b, \\
    \partial_t(hv)
    + \partial_y\left(v^2h + \frac{1}{2}g_rh^2\right)
    + \partial_x(uvh)
    &= -g_rh\,\partial_y b,
\end{aligned}
\end{equation}
where $(t,x,y)\in[0,T]\times\Omega$, $h$ denotes the water depth, $u$ and $v$ are the velocity components,
$b$ is the spatially varying bathymetry, and $g_r$ denotes the gravitational
acceleration. The quantities $hu$ and $hv$ represent the directional momentum
components. 

We use the publicly available SWE benchmark from the PDEBench repository, which considers a two-dimensional radial dam-break problem on the square domain $\Omega=[-2.5,2.5]^2$. The initial water depth is given by
\begin{equation}
h(0,x,y)=
\begin{cases}
2, & \sqrt{x^2+y^2}<r,\\
1, & \sqrt{x^2+y^2}\ge r,
\end{cases}
\qquad r\sim\mathcal{U}(0.3,0.7).
\end{equation}
The benchmark is provided at a spatial resolution of $128\times128$. We perform experiments at resolution $64\times64$, with the time $T\!=\!101$.

\paragraph{3D Rayleigh--Taylor Instability.} The Rayleigh--Taylor instability (RTI) describes the unstable evolution of two fluids with different densities under acceleration, where perturbations at the interface grow into characteristic rising bubbles and falling spikes and eventually develop into nonlinear turbulent mixing. A key dimensionless parameter is the Atwood number $A_t$,
\[
A=\frac{\rho_h-\rho_l}{\rho_h+\rho_l},
\]
where $\rho_h$ and $\rho_l$ denote the densities of the heavy and light fluids, respectively. The miscible RTI system considered in The Well benchmark is governed by
\begin{align}
    \partial_t \rho + \nabla\cdot(\rho\mathbf{u}) &= 0,\\
    \partial_t(\rho\mathbf{u}) + \nabla\cdot(\rho\mathbf{u}\mathbf{u})
    &= -\nabla p + \nabla\cdot\boldsymbol{\tau} + \rho\mathbf{g},\\
    \nabla\cdot\mathbf{u}
    &= -\kappa\nabla\cdot\left(\frac{\nabla\rho}{\rho}\right),
\end{align}
where $\rho$ is the density, $\mathbf{u}$ is the velocity field, $p$ is the pressure, $\mathbf{g}$ denotes gravity, and $\kappa$ is the molecular diffusivity. The deviatoric stress tensor is
\[
\boldsymbol{\tau}
=
\rho\nu
\left(
\nabla\mathbf{u}
+
(\nabla\mathbf{u})^{\top}
-
\frac{2}{3}(\nabla\cdot\mathbf{u})\mathbf{I}
\right),
\]
with $\nu$ denoting the kinematic viscosity.
We use the publicly available 3D RTI dataset from The Well~\cite{ohana2024well}, which is originally provided on a $128\times128\times128$ grid and $A_t=0.25$ with 120 temporal snapshots per trajectory. Due to the substantially higher memory cost of three-dimensional operator learning, we spatially downsample each snapshot to $32\times32\times32$ for all experiments. We further retain the first 100 simulation steps and temporally subsample the trajectories by retaining one snapshot every four steps, resulting in sequences of 25 states. While the original dataset contains both density and velocity fields, we consider only the scalar density field $\rho(\mathbf{x},t)$ as the predicted state in this work.

\section{Baseline Methods}\label{app:baseline}
In this section, we describe details about the baseline methods considered in this study. 

\subsection{Rollout Baselines}

\textbf{Fourier Neural Operator (FNO)}~\cite{li2020fourier}:
FNO learns a resolution-invariant mapping between function spaces by (i) lifting the input field to a latent channel space, (ii) applying a stack of Fourier operator blocks that combine a non-local spectral operator with local pointwise maps and nonlinearities, and (iii) projecting back to the solution space. A single FNO block can be written as
\begin{equation}
v^{(\ell+1)}(x) = \sigma \Big( W^{(\ell)} v^{(\ell)}(x) + \mathcal{K}^{(\ell)}(\phi) v^{(\ell)}(x) \Big),
\label{eq:fno}
\end{equation}
where $\mathcal{K}^{(\ell)}$ denotes the learned kernel integral operator parameterized by $\phi$, $W^{(\ell)}$ is a local linear map, and $\sigma$ is a pointwise nonlinearity.


\textbf{Factorized Fourier Neural Operators (FFNOs)}~\cite{tran2021factorized}: 
To address computational bottlenecks in high-dimensional settings and enable the training of deeper architectures, the factorized Fourier neural operator (FFNO)~\cite{tran2021factorized} introduces a separable spectral layer. Instead of a full $d$-dimensional Fourier transform, FFNO factorizes the operation across each spatial dimension independently. The kernel operator is decomposed into a sum of spectral convolutions over dimensions $j \in \{1, \dots, d\}$:
\begin{equation}
    \mathcal{K}^{(\ell)}_{\text{fac}}(\phi) v^{(\ell)}(x) = \sum_{j=1}^d \mathcal{F}_j^{-1} \Big( R_{\phi, j} \cdot \mathcal{F}_j(v^{(\ell)}) \Big)(x),
    \label{eq:ffno_kernel}
\end{equation}
where $\mathcal{F}_j$ is the 1D Fourier transform along dimension $j$. Furthermore, FFNO adopts an enhanced residual block structure to improve gradient flow. The update rule for layer $l$ incorporates a two-layer feedforward network (MLP) mixed with the spectral convolution:
\begin{equation}
    v^{(\ell+1)}(x) = v^{(\ell)}(x) + \sigma \left( W_2^{(\ell)} \sigma \left( W_1^{(\ell)} (\mathcal{K}^{(\ell)}_{\text{fac}}(\phi) v^{(\ell)}(x)) + b_1^{(\ell)} \right) + b_2^{(\ell)} \right),
\end{equation}

where $\{W_{i}\}$ are local linear transformations and $\{b_{i}\}$ are biases. This format significantly reduces the parameter count and computational complexity while maintaining the global receptive field.


\textbf{U-Net Neural Operator (U-Net)}~\cite{gupta2022towards}: As a convolutional encoder-decoder baseline, we use a U-Net style  with skip connections. It consists of four downsampling convolution blocks, a middle bottleneck block, and four upsampling convolution blocks. The upsampling path uses the standard U-Net design with skip connections between corresponding stages in the encoder and decoder. We use channel multipliers [1,2,2,2], do not employ explicit time embeddings, and set the first dimension to 32. 

\textbf{S4 Factorized Fourier Neural Operator (S4FFNO)}~\cite{buitrago2025benefits}: S4FFNO is a memory-augmented model that instantiates the MemNO framework by combining FFNO as a Markovian neural operator with an explicit sequential memory layer based on S4 \cite{gu2021efficiently}. Concretely, the spatial operator layers are applied timestep-wise as in FFNO, while memory is introduced by inserting a causal sequence model $M$ on a chosen hidden layer to aggregate information from past states. In S4FFNO, this memory module is implemented with an S4 layer operating independently at each spatial location $x$, i.e., 
\begin{equation}
\tilde v^{(k)}(t_i,x)
\;=\;
\mathrm{S4}\Big( v^{(k)}(t_i,x),\,v^{(k)}(t_{i-1},x),\,\ldots,\,v^{(k)}(t_0,x)\Big),
\label{eq:s4ffno_memory}
\end{equation}
and the resulting $\tilde v^{(k)}(t_i,\cdot)$ is then passed through the remaining FFNO layers.

\textbf{PDE-Refiner}~\cite{lippe2023pde}: PDE-Refiner improves autoregressive neural PDE solvers by repeatedly refining the prediction of the same next state. The main motivation is that standard one-step objectives tend to emphasize dominant, high-amplitude frequency components, while low-amplitude components may be underrepresented and accumulate into substantial errors during long rollouts.

Following the diffusion-based formulation of PDE-Refiner, we predict the temporal residual rather than the next state directly. Given the current state $u_t$, we define the normalized residual as
\begin{equation}
    r_{t+1}
    =
    \frac{u_{t+1}-u_t}{c_{\mathrm{out}}},
\end{equation}
where $c_{\mathrm{out}}$ is a dataset-dependent output factor used to normalize the residual scale. Let $\beta_s$ denote the noise variance at diffusion level $s$, with $\alpha_s=1-\beta_s$ and $\bar{\alpha}_s=\prod_{j=0}^{s}\alpha_j$. During training, a noise level $s$ is sampled and Gaussian noise is added to the normalized residual:
\begin{equation}
    \tilde r_{t+1}^{(s)}
    =
    \sqrt{\bar{\alpha}_s}\,r_{t+1}
    +
    \sqrt{1-\bar{\alpha}_s}\,\epsilon,
    \qquad
    \epsilon\sim\mathcal{N}(0,I).
\end{equation}
The neural operator takes $u_t$, the noisy residual candidate $\tilde r_{t+1}^{(s)}$, and the noise level $s$ as input. We adopt the diffusion-based $v$-prediction objective, with target
\begin{equation}
    v^{(s)}
    =
    \sqrt{\bar{\alpha}_s}\,\epsilon
    -
    \sqrt{1-\bar{\alpha}_s}\,r_{t+1}.
\end{equation}
At inference time, the reverse process iteratively transforms the candidate from the highest noise level $s=K$ to the lowest level $s=0$, yielding the refined residual prediction $\hat r_{t+1}$. The next state is reconstructed as $\hat u_{t+1}=u_t+c_{\mathrm{out}}\hat r_{t+1}$ and then propagated autoregressively to generate the full trajectory.

For PDE-Refiner, we use the same FNO backbone and dataset-specific backbone configurations described in Section~\ref{subsec:model_specification}, so that the comparison differs only in the refinement mechanism rather than model capacity. We use three refinement steps and the exponential noise schedule proposed in PDE-Refiner~\cite{lippe2023pde}. For the minimum noise level, we follow the original PDE-Refiner settings for KSE and NSE, using $\sigma_{\min}^2=2\times10^{-7}$ for KSE and $\sigma_{\min}^2=10^{-3}$ for NSE. For BE and SWE, which are not considered in the original PDE-Refiner study, we transfer the setting from the benchmark with the same spatial resolution, using $\sigma_{\min}^2=2\times10^{-7}$ for BE, consistent with KSE, and $\sigma_{\min}^2=10^{-3}$ for SWE, consistent with NSE. For RTI, we use $\sigma_{\min}^2=4\times10^{-7}$. The output factor $c_{\mathrm{out}}$ is set according to the standard deviation of the temporal residual for each dataset. Specifically, for KSE with $\nu\in\{0.075,0.1,0.125\}$, we use $c_{\mathrm{out}}\in\{0.4,0.18,0.12\}$, respectively. For NSE with $\nu\in\{10^{-5},10^{-4},10^{-3}\}$, we use $c_{\mathrm{out}}\in\{0.031,0.029,0.018\}$, respectively. We set $c_{\mathrm{out}}=0.008$ for SWE, $c_{\mathrm{out}}=0.12$ for BE, and $c_{\mathrm{out}}=0.0125$ for RTI. These factors are computed from the corresponding temporal residuals at resolution $S=128$ for KSE and BE, $64\times64$ for NSE and SWE, and $32\times32\times32$ for RTI.

\textbf{FreqMoE}~\cite{chen2025freqmoe}: FreqMoE augments Fourier Neural Operators with frequency-domain mixture-of-experts to improve the modeling of high-frequency components.  It follows a low-frequency pretraining and high-frequency fine-tuning (LPHF) paradigm. A low-frequency FNO is first pretrained and then upcycled into a frequency-domain mixture-of-experts model, where the pretrained spectral weights are retained as the low-frequency base component and specialized high-frequency experts are introduced through low-rank adaptation. The frequency spectrum beyond the base band is partitioned into disjoint regions assigned to different experts, while a learnable gating mechanism controls their contributions.

For FreqMoE, we adopt the same FNO backbone and dataset-specific configurations as our FNO baselines. Following the LPHF paradigm, the low-frequency FNO is pretrained for 100 epochs and then sparsely upcycled from its final checkpoint for 100 epochs of high-frequency fine-tuning, preserving the same total 200-epoch training budget as baselines. For BE and KSE at spatial resolution $S=128$, the pretrained FNO uses 48 Fourier modes and the high-frequency experts extend the spectral support to all 65 nonnegative Fourier modes. For NSE and SWE at resolution $64\times64$, the pretrained FNO uses 16 modes per spatial dimension and sparse upcycling extends the spectral support to $(32,33)$, corresponding to the full real-valued Fourier spectrum. For RTI at resolution $32\times32\times32$, the pretrained FNO uses 12 modes per spatial dimension and sparse upcycling extends the spectral support to $(16,16,17)$, corresponding to the full real-valued Fourier spectrum. We use LoRA~\cite{hu2021lora} rank 4, a sparsity regularization weight of 0.01, and Top-1, Top-2, and Top-5 expert activation for the 1D, 2D, and 3D benchmarks, respectively. Both the pretraining and fine-tuning stages use one-step teacher forcing, while evaluation is performed through fully autoregressive rollout.

\subsection{Flow-map Baselines}

\textbf{HyperFNO}~\cite{alesiani2022hyperfno}: HyperFNO is a Fourier Neural Operator whose weights are conditioned by a small hyper-network. Different from the original setting where the hyper-network is conditioned on the PDE parameter vector $\mu$, we condition it only on the prediction time increment $\Delta_t$. Concretely, for each training sample, we feed $\Delta_t$ to a 3-layer MLP hyper-network with width 64 to generate a conditioning vector. This vector is then used to produce $\Delta_t$-dependent adjustments to the FNO parameters, including the lift/projection operators, the spectral convolution weights, and pointwise weights. For each Fourier layer $l$, the hyper-network produces additive updates to the spectral convolution weights and pointwise linear map.  For the spectral convolution, we start from base complex weights $R_0$ and add a low-rank, $\Delta_t$-dependent factor scaled by a learned complex tensor $R_1$: $R(\Delta_t) = R_0 + \Delta R(\Delta_t)\odot R_1,$
where $\Delta R(\Delta_t)$ is constructed via a low-rank factorization driven by the conditioning vector. Similarly, the pointwise map uses an additive form $W(\Delta_t) = W_0 + \Delta W(\Delta_t)W_1,$ where the row-wise scaling vector $\Delta W(\Delta_t)$ is obtained by linearly projecting the conditioning vector. This yields a time-conditioned operator that can be queried at arbitrary $\Delta_t$ while keeping the conditioning overhead small.

\textbf{MLP}: This Multi-Layer Perceptron (MLP) baseline operates directly in the physical domain. The architecture follows a Lifting-Processor-Projection  pipeline designed to learn the flow-map $F_{s,t}(u_s) = u_t$ by fusing local and global information into a unified representation. Specifically, the initial state $u_s$, the spatial grid coordinates, and the temporal displacement $(t-s)$ are concatenated into a single input vector, providing the network with the simultaneous access to the local state and the global time coordinate. This fused input is mapped to a latent space via a linear lifting layer, followed by four sequential residual blocks defined as $v^{(l+1)} = v^{(l)} + \sigma(W^{(l)}v^{(l)} + b^{(l)})$. Finally, a linear projection layer maps the latent features back to the physical space to produce the predicted field.

\textbf{DeepONet (DON)}~\cite{lu2019deeponet}: DeepONet approximates an operator using two subnetworks, a branch network that encodes the input function and a trunk network that encodes the query location. Predictions are obtained by combining branch and trunk features via an inner product style aggregation.


\textbf{Time-conditioned InstanceNorm (FNO-IN)}~\cite{herde2024poseidon}: To incorporate variable lead times into Fourier-based neural operators, we adopt lead-time conditioned instance normalization as a conditioning mechanism. Given an intermediate feature field $v(x)$, FNO-IN modulates standard instance normalization via a time-dependent affine transform: 
\begin{equation}
\mathrm{IN}_{\Delta_t}(v)(x)
\;=\;
\alpha(\Delta_t)\odot \mathrm{IN}(v)(x) \;+\; \beta(\Delta_t),
\label{eq:tcin}
\end{equation}
where $\odot$ denotes element-wise multiplication, $\mathrm{IN}(\cdot)$ is applied channel-wise, and the per-channel scale and shift $\alpha(\Delta_t),\beta(\Delta_t)\in\mathbb{R}^{C}$ are produced by a small MLP that takes the lead time $\Delta_t$ as input and outputs per-channel modulation parameters. We apply IN inside each operator layer, enabling a single model to condition its dynamics on arbitrary time increments. For a FNO-style block, the FNO-IN update can be written as 
\begin{equation}
v^{(\ell+1)}(x)
=
\sigma\!\left(
\mathrm{IN}_{\Delta_t}\!\Big(
W^{(\ell)} v^{(\ell)}(x) + \mathcal{K}^{(\ell)}(v^{(\ell)})(x)
\Big)\right),
\label{eq:fno_tcin_layer}
\end{equation}
where $W^{(\ell)}$ is a pointwise linear map, $\mathcal{K}^{(\ell)}$ is the spectral (Fourier) convolution operator, and $\sigma$ is a pointwise nonlinearity.

A FFNO-style block can be written as 
\begin{equation}
\begin{aligned}
z^{(\ell)}(x) &= W_1^{(\ell)}\big((\mathcal{K}_{\mathrm{fac}}(\phi)\,v^{(\ell)})(x)\big)+b_1^{(\ell)},\\
v^{(\ell+1)}(x) &= v^{(\ell)}(x) + W_2^{(\ell)}\,\sigma\!\Big(\mathrm{TCIN}_{\Delta_t}\big(z^\ell(x)\big)\Big) + b_2^{(\ell)} \\
&= v^{(\ell)}(x) 
+ W_2^{(\ell)}\,\sigma\!\Big(\alpha(\Delta_t)\odot \mathrm{IN}\!\big(z^{(\ell)}(x)\big)+\beta(\Delta_t)\Big) + b_2^{(\ell)}.
\end{aligned}
\label{eq:ffno_tcin_layer}
\end{equation} 
where $W_1^{(\ell)},W_2^{(\ell)}$ and $b_1^{(\ell)},b_2^{(\ell)}$ are learned pointwise parameters and $\mathcal{K}_{\mathrm{fac}^{(\ell)}}$ denotes the factorized spectral operator.

\section{Implementation Details}
\label{app:implementation}
In this section, we provide a detailed description of our training strategies for autoregressive rollouts, model specifications, and the hardware configurations used in our experiments.

\subsection{Training Strategies for Autoregressive Rollouts}
Let $\{u^t\}_{t=0}^{T_{total}}$ denote a discrete-time trajectory and let $\mathcal{M}$ denote a one-step predictor. 

\textbf{Rollout (autoregressive)}: In rollout setting, the model is trained and evaluated by feeding back its own predicted states as inputs for subsequent steps: 
\begin{equation}
\hat{u}^{t+1} = \mathcal{M}(\hat{u}^{t}), \qquad \hat{u}^{0}=u^{0}.
\end{equation}

\textbf{Markov teacher forcing (TF)}: In teacher forcing, the model is trained with the ground-truth previous state as input, while the model is evaluated autoregressively:
\begin{equation}
\label{eq:TF_trainVStest}
\hat{u}^{t+1} = \mathcal{M}(u^{t}) \quad \text{(train)}, \qquad
\hat{u}^{t+1} = \mathcal{M}(\hat{u}^{t}) \quad \text{(test)}.
\end{equation}
We use a first-order (Markov) conditioning scheme, i.e., the model conditions only on the immediately preceding state.

\subsection{Training Strategies for Hybrid Execution Modes}
\paragraph{Two training modes.} 
The segmented flow-map framework introduced in Section~\ref{sec:hybrid} admits two operational modes during training. In Segmented Rollout (Hybrid-RO), the predicted terminal state of each segment is recursively fed into the next segment, enabling fully autonomous inference. In Segmented Teacher Forcing (Hybrid-TF), the model is re-grounded using ground-truth states at segment boundaries during training, which stabilizes optimization by preventing early prediction errors from corrupting subsequent segments. The distinction is made explicit in Algorithm~\ref{alg:hybrid}. 

\paragraph{Limiting cases and trade-offs.} 
The segment length $\tau$ interpolates between two extremes:
\begin{equation}
\begin{split}
\tau &= 1 \;\; \Rightarrow\;\; \text{standard autoregressive rollout}, \\
\tau &= T \;\; \Rightarrow \;\; \text{single global flow-map}.    
\end{split}
\end{equation}
Intermediate values of $\tau$ define a continuum of execution regimes that balance local stability and global efficiency. Smaller segments reduce the difficulty of learning long-range mappings at the cost of more frequent recursion, while larger segments improve computational efficiency but increase the burden on the flow-map operator.

By explicitly bridging rollouts and flow-map at the operational level, segmented execution provides a flexible mechanism for stabilizing long-horizon prediction without sacrificing the core advantages of the flow-map prediction formulation.

\begin{algorithm}[t]
\caption{Hybrid Segmented Flow-map Training}
\label{alg:hybrid}
\begin{algorithmic}
\STATE {\bfseries Input:} Training trajectory $\{u^t\}_{t=0}^{T_{total}}$, segment length $\tau$, mode $m\in\{\text{Hybrid-RO},\text{Hybrid-TF}\}$, model $\hat{F}_\theta(\cdot,\Delta_t)$
\STATE {\bfseries Define:} $T := T_{total}-1$, number of segments $K := \left\lceil \frac{T}{\tau}\right\rceil$
\STATE Initialize total loss $L \leftarrow 0$
\STATE Initialize segment start index $s \leftarrow 0$
\STATE Initialize segment input state $w \leftarrow u^0$

\FOR{$j = 1$ {\bfseries to} $K$}
    \STATE $r \leftarrow \min(\tau,\; T-s)$ 
    \STATE Compute segment predictions (direct flow-map jumps):
    \FOR{$\Delta_t = 1$ {\bfseries to} $r$}
        \STATE $\hat{u}^{s+\Delta_t} \leftarrow \hat{F}_\theta(w,\Delta_t)$
        \STATE $L \leftarrow L + \dfrac{\left\|\hat{u}^{s+\Delta_t}-u^{s+\Delta_t}\right\|_2}{\left\|u^{s+\Delta_t}\right\|_2}$ 
    \ENDFOR

    \STATE Update for next segment boundary:
    \IF{$m=\text{Hybrid-RO}$}
        \STATE Set boundary state $\hat{u}_{end} \leftarrow \hat{u}^{s+r}$ 
        \STATE Reset input state from predictions: \STATE $w \leftarrow \hat{u}_{end}$

    \ELSIF{$m=\text{Hybrid-TF}$}
        \STATE Reset window from ground truth: $w \leftarrow u^{s+r}$
    \ENDIF
    \STATE $s \leftarrow s + r$
\ENDFOR

\STATE $L \leftarrow \dfrac{1}{T}\,L$
\STATE {\bfseries Output:} $L$ (use to update $\theta$ via backprop)
\end{algorithmic}
\end{algorithm}

\subsection{Training Details}
 All methods are implemented in Python using PyTorch~\cite{paszke2019pytorch}, building on the original implementation of FNOs~\cite{li2020fourier}, and following the experimental setup in \cite{buitrago2025benefits}. 
 
 All models are trained for 200 epochs using the AdamW optimizer~\cite{loshchilov2017decoupled} with an initial learning rate of 0.001, weight decay 0.0001, and a cosine-annealing learning-rate schedule. The batch sizes are set to 32 for KSE, 64 for BE, 20 for NSE, 10 for SWE, and 1 for RTI. For each configuration, we run three independent experiments with varying random seeds and report results averaged over three runs of the test loss at the final training epoch. Experiments are conducted on Nvidia RTX 4090 and A100 GPUs in a single-GPU setup. Although we do not optimize our code particularly to minimize the inference time, the proposed FiLM-only baselines naturally take less time per epoch to train compared to rollout-based methods. For our standard NSE setup, baselines with only FiLM module take around 40 seconds per epoch, while the rollout-based methods take roughly 120 seconds per epoch.

\subsection{Model Specification}
\label{subsec:model_specification}
For all FNO backbones, we use a unified architecture and hyper-parameter configuration to control for model capacity. Specifically, we employ four Fourier layers. For KSE and BE, all non-frequency-decomposed models use the full set of spectral modes, i.e., $k_{max} = \lfloor S/2 \rfloor$ (where $S$ is the spatial resolution). For NSE, we follow the setting used in prior baselines~\cite{tran2021factorized, buitrago2025benefits} to use 16 Fourier modes. For SWE, we use 16 Fourier modes. For RTI, we use 12 Fourier modes. We set the width to $d_v = 128$ for KSE and BE, while for NSE, SWE, and RTI, we use $d_v = 64$. The input is lifted to $d_v$ channels by a linear layer, and the output is projected back to solution space using a two-layer MLP with one hidden layer, the hidden dimension is 256 for KSE and BE, 128 for NSE, SWE, and RTI. The expanded hidden dimension of the feedforward MLP of FFNO is $4 \cdot d_v$. The nonlinearity functions are GELU~\cite{hendrycks2016gaussian} for KSE and BE, and RELU~\cite{nair2010rectified} for NSE, SWE, and RTI. 

For our proposed time-conditioned variants, we use a single-layer MLP with width $d_v$ to generate the FiLM modulation parameters $\gamma(\Delta_t)$ and $\beta(\Delta_t)$, where $\Delta_t$ is normalized by the total prediction horizon or by the segment length \texttt{seg} in the segmented execution setting. This FiLM hypernetwork is trained with a learning rate that is $10\times$ larger than the backbone learning rate, while all other optimizer settings remain unchanged. 

\section{
Frequency-Decomposed Spectral Refinement}
\label{app:freq}
\textbf{Frequency-based decomposition}: 
We first compute the Fourier transform of the hidden representation $v^{(\ell)}$ and truncate the resulting modes using a cutoff $m$, leading to the decomposition: 
\begin{equation}
\begin{aligned}
\widehat v^{(\ell)}
&=
\mathcal{F}\left(v^{(\ell)}\right), \quad 
\widehat v_{\mathrm{low}}^{(\ell)}
&=
M_{\leq m} \odot \widehat v^{(\ell)}, \quad
\widehat v_{\mathrm{high}}^{(\ell)}
&=
\left(1 - M_{\leq m}\right)
\odot
\widehat v^{(\ell)}.
\end{aligned}
\end{equation}
where $M_{\leq m}$ is a binary low-pass mask that keeps the first $m$ Fourier modes. A reconstruction of the hidden representations via inverse Fourier transform after applying the low- and high-frequency filters gives:
\begin{equation}
    v_{\mathrm{low}}^{(\ell)}
    =
    \mathcal{F}^{-1}
    \left(
    \widehat v_{\mathrm{low}}^{(\ell)}
    \right), \quad \text{and} \quad
    v_{\mathrm{high}}^{(\ell)}
    =
    \mathcal{F}^{-1}
    \left(
    \widehat v_{\mathrm{high}}^{(\ell)}
    \right).
\end{equation}

\textbf{Low-frequency signal processing}, $v^{(\ell)}_{\text{low}} \mapsto \hat{v}^{(\ell)}_{\text{low}}$: The low-frequency component $v_{\mathrm{low}}^{(\ell)}$ is passed through the spectral operator of the underlying backbone:
\begin{equation}
    \hat{v}^{(\ell)}_{\text{low}}=\mathcal{K}^{(\ell)}(\phi) v^{(\ell)}_{\text{low}}(x),
\end{equation}
where $\mathcal{K}$ denotes the FNO kernels.

\textbf{High-frequency signal processing}, $ \left({v}^{(\ell)}_{\text{high}}, \hat{v}^{(\ell)}_{\text{low}}\right) \mapsto \hat {v}^{(\ell)}_{\text{high}}$: We process the high-frequency component by conditioning it locally. To this end, we choose $1\times 1$ convolutional operations to maintain the discretization-invariance property and provide computational efficiency. 

Specifically, we form 
$z_{\mathrm{high}}^{(\ell)}
    =
    \left[
    v_{\mathrm{high}}^{(\ell)},
    \hat{v}_{\mathrm{low}}^{(\ell)}
    \right],
$
by concatenating them along the channel dimension. 
Then the two-layer $1\times 1$ convolutional neural network is applied 
\begin{equation}
r_{\mathrm{high}}^{(\ell)}
    =
        A_2^{(\ell)}
    \sigma
    \left(
    A_1^{(\ell)}
    z_{\mathrm{high}}^{(\ell)}
    \right),  
\end{equation}
where $A_1^{(\ell)}$ and $A_2^{(\ell)}$ are $1 \times 1$ convolution maps and $\sigma$ is a point-wise nonlinearity. Effectively, this is a local operator mixing the information between the low-frequency component and the high-frequency component at each spatial location; that is, at $x$, the $1\times1$ convolution performs weighted summation such that $\kappa_1 v_{\text{high}}(x) + \kappa_2 \hat v_{\text{low}}(x) + b$, where $\kappa_i$ are convolutional kernel weights and $b$ denote the bias term. 

\textbf{Time-displacement conditioning}, $ \left(\hat{v}^{(\ell)}_{\text{low}}, \hat{v}^{(\ell)}_{\text{high}}\right) \mapsto \tilde {v}^{(\ell+1)}$: The low-frequency spectral term and the high-frequency refinement term are combined using a time-conditioned fusion gate. For each layer, we introduce a small MLP $\rho^{(\ell)} : \mathbb{R}^{1} \to \mathbb{R}^{2}$, which maps the conditioning feature $\Delta t$ to two fusion
coefficients:$
    \left[
    g_{\mathrm{low}}^{(\ell)}(\Delta_t),
    g_{\mathrm{high}}^{(\ell)}(\Delta_t)
    \right]
    =
    \rho^{(\ell)}(\Delta_t).$
After inferring the coefficients, the processed components are combined as a weighted linear combination: 
\begin{equation}
    \tilde v^{(\ell+1)}
    =
    g_{\mathrm{low}}^{(\ell)}(\Delta_t)
    \odot
    r_{\mathrm{low}}^{(\ell)}
    +
    g_{\mathrm{high}}^{(\ell)}(\Delta_t)
    \odot
    r_{\mathrm{high}}^{(\ell)}.
\end{equation}
This allows the operator to vary the relative contribution of low-frequency spectral evolution and high-frequency local refinement according to the target time displacement.

\subsection{Frequency-Dependent Evolution and Prediction Errors}
\begin{figure}[t]
    \centering
    \includegraphics[width=0.48\linewidth]{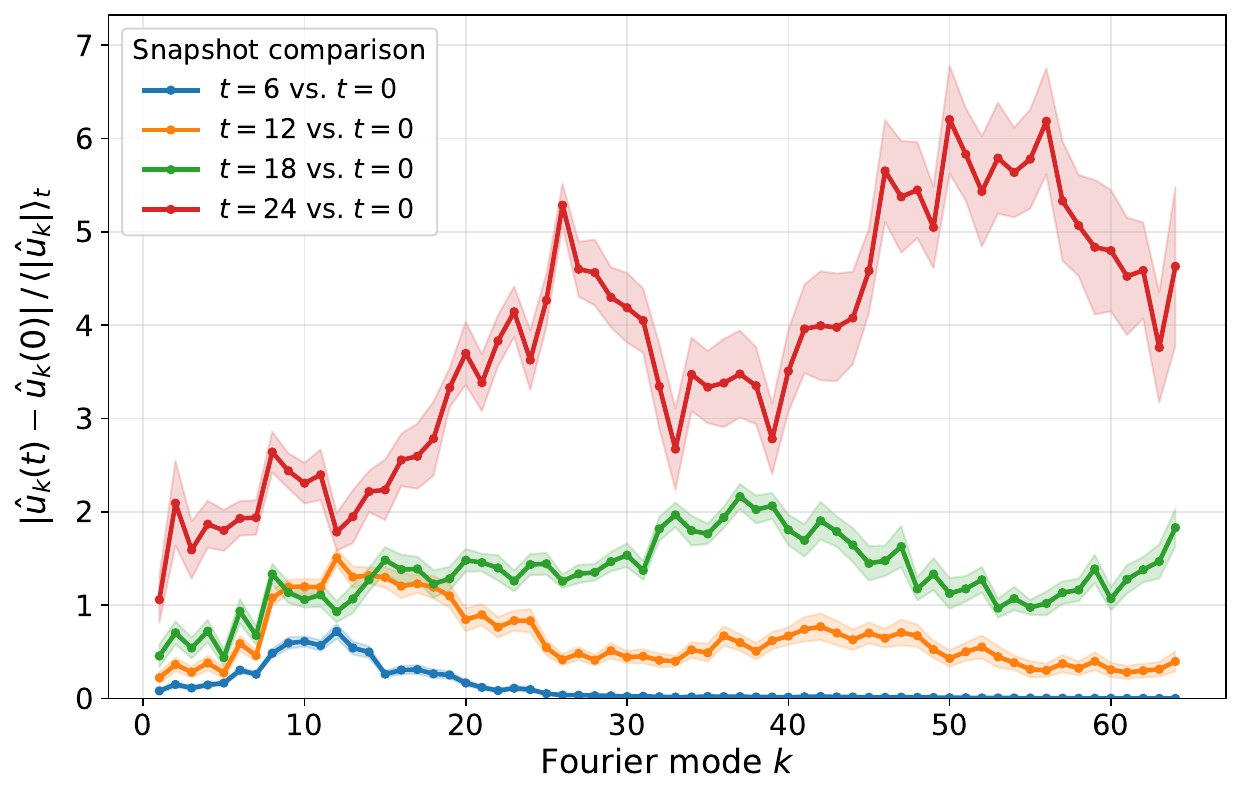}
    \vspace{-0.8em}
    \caption{[KSE, $\nu=0.075$, $S=128$] Per-frequency Fourier coefficient change from the initial state.}
    \label{fig:freq_change}
    \vspace{-1em}
\end{figure}

\begin{figure}[b]
    \centering
    \vspace{-1em}
    \includegraphics[width=0.48\linewidth]{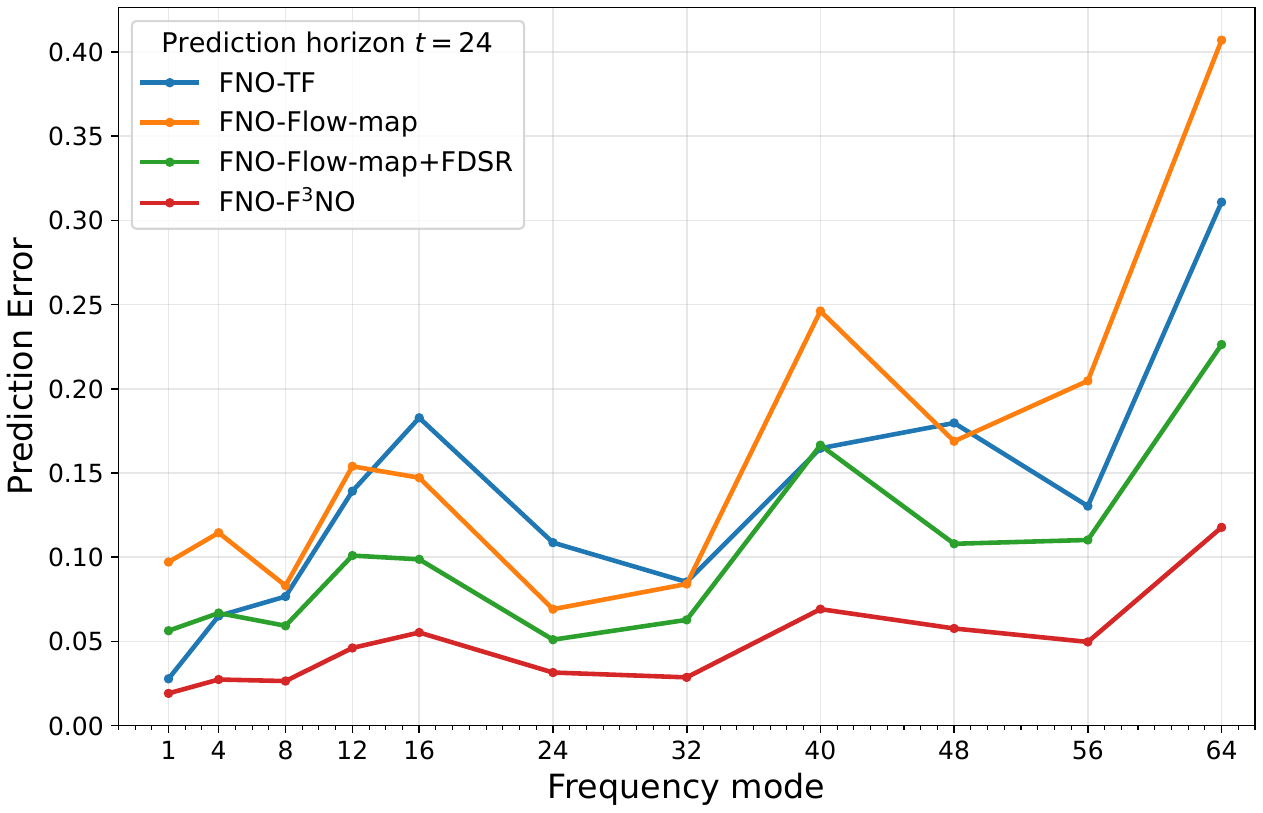}
    \vspace{-0.8em}
    \caption{[KSE, $\nu = 0.075$, $S = 128$] Mean per-mode relative $L^2$ error as a function of Fourier mode $k$ at the final prediction step $t=24$. The curves correspond to the final column of each heatmap in Figure~\ref{fig:freqloss} and report time-instantaneous (non-cumulative) errors.}
    \label{fig:per_mode_error_h24}
\end{figure}

Figure~\ref{fig:freq_change} visualizes the per-frequency evolution of the KSE trajectories in Fourier space from the initial state at resolution $S=128$ and viscosity $\nu=0.075$. For each Fourier mode $k$, we report the relative change $|\hat{u}_k(t)-\hat{u}_k(0)| / \langle |\hat{u}_k| \rangle_t$ 
at four future snapshots, $t=6,12,18,24$, each compared against $t=0$. For each time step, the relative change is averaged over 20 samples. The result indicates that the low frequency remains relatively stable (i.e., less changes), especially at early time steps, while the dominant changes grow over time and become concentrated in the mid-to-high frequency range. Moreover, as the time displacement becomes larger ($t=18$ and $t=24$), the changes in high-frequency are more pronounced. These findings suggest that different spectral components exhibit different temporal evolution patterns, motivating a frequency-aware mechanism rather than treating all modes together. 

Figure~\ref{fig:per_mode_error_h24} examines the frequency-resolved errors at the final prediction step, corresponding to the last column of each heatmap in Figure~\ref{fig:freqloss}. FNO-Flow-map does not consistently improve upon FNO-TF, and both retain substantial errors at high Fourier modes, showing that direct prediction alone is insufficient to address these errors. Adding FDSR to the direct flow-map predictor reduces the error at every displayed mode, although appreciable errors remain at the highest frequencies. With segmented training and execution, F\textsuperscript{3}NO further suppresses high-frequency errors, achieving the lowest error across all displayed modes and the most consistent error curve. These results support the complementary contributions of spectral refinement and segmentation to long-horizon prediction accuracy.

\begin{table}[t]
  \caption{[KSE, $S = 128$] Relative $L^2$ error (\%) with mean $\pm$ standard deviation for varying $\nu$ to isolate the effect of low-frequency conditioning and nonlinear local branch. We isolate the effect of low-frequency conditioning by comparing the high-frequency branch that uses high-frequency features only, high-frequency features concatenated with the original low-frequency feature, and high-frequency features concatenated with the spectrally updated low-frequency feature used in FDSR. We also replace the high-frequency branch in FDSR with a parameter-matched pointwise MLP applied to the full undecomposed latent feature, while keeping the low-frequency spectral branch and all remaining components unchanged.}
  \label{tab:kse_lowfreq_ablation128}
  \centering
      \scriptsize
    \setlength{\tabcolsep}{4pt}
  \begin{tabular}{lccccc}
    \toprule
    \multirow{2}{*}{Model} & \multirow{2}{*}{Modes} & \multirow{2}{*}{Params.} & \multicolumn{3}{c}{$\nu$}                 \\
    \cmidrule(r){4-6}
        &  &  & 0.075 & 0.1 & 0.125 \\
    \midrule
    \multicolumn{6}{l}{Flow-map} \\
    \cmidrule(lr){2-6}
    FNO-Flow-map & 65 & 4,361,601 & 3.38$\pm$0.06 & 0.90$\pm$0.04 & 0.43$\pm$0.00 \\
    FNO-Flow-map+FDSR & 48 & 3,447,177 & 2.01$\pm$0.05 & 0.63$\pm$0.01 & 0.34$\pm$0.00 \\
    \midrule
    \multicolumn{6}{l}{Hybrid-TF, $\tau=4$} \\
    \cmidrule(lr){2-6}
    FNO-Flow-map+FDSR (High-freq only) & 48 & 3,447,177 &1.25$\pm$0.05 & 0.36$\pm$0.01 & 0.19$\pm$0.00\\
    FNO-Flow-map+FDSR (High-freq concat with original low-freq) & 48 & 3,447,177 &1.24$\pm$0.06 & 0.34$\pm$0.03 & 0.18$\pm$0.00\\
    FNO-Flow-map+FDSR (MLP local branch) & 48 & 3,447,473 &1.25$\pm$0.03 & 0.37$\pm$0.01 & 0.19$\pm$0.00\\
    FNO-Flow-map+FDSR (F\textsuperscript{3}NO) & 48 & 3,447,177 & \textbf{1.15$\pm$0.06} &\textbf{0.32$\pm$0.00} & \textbf{0.17$\pm$0.01}\\
    
    \bottomrule
  \end{tabular}
\end{table}

\begin{table}[b]
  \caption{[NSE, $S = 64\times64$] Relative $L^2$ error (\%) with mean $\pm$ standard deviation for varying $\nu$ to isolate the effect of low-frequency conditioning and nonlinear local branch. We isolate the effect of low-frequency conditioning by comparing the high-frequency branch that uses high-frequency features only, high-frequency features concatenated with the original low-frequency feature, and high-frequency features concatenated with the spectrally updated low-frequency feature used in FDSR. We also replace the high-frequency branch in FDSR with a parameter-matched pointwise MLP applied to the full undecomposed latent feature, while keeping the low-frequency spectral branch and all remaining components unchanged.}
  \label{tab:nse_lowfreq_ablation128}
  \centering
      \scriptsize
    \setlength{\tabcolsep}{4pt}
  \begin{tabular}{lccccc}
    \toprule
    \multirow{2}{*}{Model} & \multirow{2}{*}{Modes} & \multirow{2}{*}{Params.} & \multicolumn{3}{c}{$\nu$}                 \\
    \cmidrule(r){4-6}
        &  &  & $\nu=10^{-5}$ & $\nu=10^{-4}$ & $\nu=10^{-3}$ \\
    \midrule
    \multicolumn{6}{l}{Flow-map} \\
    \cmidrule(lr){2-6}
    FNO-Flow-map & 16 & 8,414,977 & 5.77$\pm$0.36 & 4.90$\pm$0.09 & 1.22$\pm$0.07 \\
    FNO-Flow-map+FDSR & 16 & 8,466,697 & 4.88$\pm$0.11 & 4.79$\pm$0.50 & 1.08$\pm$0.08 \\
    \midrule
    \multicolumn{6}{l}{Hybrid-TF, $\tau=4$} \\
    \cmidrule(lr){2-6}
    FNO-Flow-map+FDSR (High-freq only) &16 & 8,466,697  & 1.67$\pm$0.00 & 0.61$\pm$0.02 & 0.26$\pm$0.02 \\
    FNO-Flow-map+FDSR (High-freq concat with original low-freq) & 16 & 8,466,697 & 1.57$\pm$0.03 & 0.59$\pm$0.00 & 0.25$\pm$0.02 \\
    FNO-Flow-map+FDSR (MLP local branch) & 16 & 8,466,825 & 1.68$\pm$0.01 & 0.92$\pm$0.02 & 0.24$\pm$0.01 \\
    FNO-Flow-map+FDSR (F\textsuperscript{3}NO) & 16 & 8,466,697 & \textbf{1.44$\pm$0.04} & \textbf{0.58$\pm$0.02} & \textbf{0.22$\pm$0.00}\\
    
    \bottomrule
  \end{tabular}
\end{table}

\subsection{Cross-Scale Conditioning and Local-Refinement}\label{app:ablation-FDSR}

We examine two architectural choices in FDSR through ablations on KSE (Table~\ref{tab:kse_lowfreq_ablation128}) and NSE (Table~\ref{tab:nse_lowfreq_ablation128}). First, we assess whether information from lower frequencies helps refine features at higher frequencies. We compare three versions of the refinement branch. The first version (``High-freq only") receives only the features at higher frequencies, while the second (``High-freq concat with original low-freq") also receives the original features at lower frequencies. In F\textsuperscript{3}NO, the spectral operator first updates the features at lower frequencies before passing them to the refinement branch alongside the features at higher frequencies.

Second, we test whether adding local nonlinear processing can provide the same benefit without separating the features by frequency. We replace the refinement branch with a pointwise MLP (``MLP local branch") that receives the full, undecomposed latent features. The resulting model has a comparable number of parameters. All four refinement variants retain the spectral branch for the lower frequencies and use the same time conditioning. Each uses Hybrid-TF with $\tau=4$.

Across both benchmarks, the full FDSR block achieves the lowest mean error at every tested viscosity. Providing the original features at lower frequencies improves upon refinement using only the features at higher frequencies. Updating those features through the spectral operator further reduces the mean error in every setting. These results support updating the information at lower frequencies before using it to guide refinement at higher frequencies. The MLP control also yields higher mean errors at comparable parameter counts, suggesting that additional nonlinear capacity alone does not explain the improvement.

\subsection{Spectral Cutoff, Parameter Efficiency, and Spatial Resolution}
Here we present further empirical analysis on F\textsuperscript{3}NOs and ablation on architectural design choices: FiLM and frequency-decomposed refinement block. Table~\ref{tab:kse_freq_ablation128} reports the prediction error for 1D KSE with a resolution $S=128$ and varying viscosity values.

\begin{table}[b]
  \caption{[KSE] Relative $L^2$ error (\%) with mean $\pm$ standard deviation for varying $\nu$ with Frequency-Decomposed Spectral Refinement and a fixed resolution $S=128$.}
  \label{tab:kse_freq_ablation128}
  \centering
      \scriptsize
    \setlength{\tabcolsep}{6pt}
  \begin{tabular}{lccccc}
    \toprule
    \multirow{2}{*}{Model} & \multirow{2}{*}{Modes} & \multirow{2}{*}{Number of Parameters} & \multicolumn{3}{c}{$\nu$}                 \\
    \cmidrule(r){4-6}
        &  &  & 0.075 & 0.1 & 0.125 \\
    \midrule
    \multicolumn{6}{l}{Rollout} \\
    \cmidrule(lr){2-6}
    FNO-TF &  65 & 4,359,553 & 3.64$\pm$0.08 & 1.48$\pm$0.10 & 0.80$\pm$0.04 \\
    FFNO-TF & 65 & 9,080,193 & 1.26$\pm$0.06 & 0.47$\pm$0.04 & 0.35$\pm$0.01 \\
    \midrule
    \multicolumn{6}{l}{Rollout with Frequency-Decomposed Spectral Refinement} \\
    \cmidrule(lr){2-6}
    FNO-TF &  32 & 2,394,505 & 6.18$\pm$0.06 & 1.13$\pm$0.03 & 0.51$\pm$0.04 \\
    FNO-TF &  40 & 2,918,793 & 2.40$\pm$0.10 & 0.71$\pm$0.03 & 0.38$\pm$0.03 \\
    FNO-TF &  48 & 3,443,081 & 2.22$\pm$0.08 & 0.69$\pm$0.06 & 0.37$\pm$0.01 \\
    FFNO-TF & 32 & 4,952,457 & 1.88$\pm$0.04 & 0.58$\pm$0.01 & 0.32$\pm$0.02 \\
    FFNO-TF & 40 & 6,001,033 & 1.29$\pm$0.06 & 0.44$\pm$0.01 & 0.25$\pm$0.01 \\
    FFNO-TF & 48 & 7,049,609 & 1.17$\pm$0.02 & 0.43$\pm$0.04 & 0.27$\pm$0.03 \\
    \midrule
    \multicolumn{6}{l}{Our Models with FiLM only} \\
    \cmidrule(lr){2-6}
    FNO-Flow-map & 65 & 4,361,601 & 3.38$\pm$0.06 & 0.90$\pm$0.04 & 0.43$\pm$0.00 \\
    \midrule
    \multicolumn{6}{l}{Our Models with Frequency-Decomposed Spectral Refinement \& FiLM} \\
    \cmidrule(lr){2-6}
    FNO-Flow-map & 48 & 3,447,177 & 2.01$\pm$0.05 & 0.63$\pm$0.01 & 0.34$\pm$0.00 \\
    F\textsuperscript{3}NO (ours) & 48 & 3,447,177 & \textbf{1.15$\pm$0.06} & \textbf{0.32$\pm$0.00} & \textbf{0.17$\pm$0.01} \\
    \bottomrule
  \end{tabular}
\end{table}

\begin{table}[t]
  \caption{[KSE] Relative $L^2$ error (\%) with mean $\pm$ standard deviation for varying $\nu$ with Frequency-Decomposed Spectral Refinement and a fixed resolution $S=64$.}
  \label{tab:kse_freq_ablation64}
  \centering
      \scriptsize
    \setlength{\tabcolsep}{6pt}
  \begin{tabular}{lccccc}
    \toprule
    \multirow{2}{*}{Model} & \multirow{2}{*}{Modes} & \multirow{2}{*}{Number of Parameters} & \multicolumn{3}{c}{$\nu$}                 \\
    \cmidrule(r){4-6}
        &  &  & 0.075 & 0.1 & 0.125 \\
    \midrule
    \multicolumn{6}{l}{Rollout} \\
    \cmidrule(lr){2-6}
    FNO-TF &  33 & 2,262,401 & 37.24$\pm$0.79 & 3.30$\pm$0.07 & 1.10$\pm$0.07 \\
    FFNO-TF & 33 & 4,885,889 & 13.20$\pm$0.20 & 1.55$\pm$0.07 & 0.46$\pm$0.03 \\
    \midrule
    \multicolumn{6}{l}{Rollout with Frequency-Decomposed Spectral Refinement} \\
    \cmidrule(lr){2-6}
    FNO-TF &  16 & 1,345,929 & 37.05$\pm$1.17 & 6.80$\pm$0.37 & 2.28$\pm$0.02 \\
    FNO-TF &  20 & 1,608,073 & 38.17$\pm$3.00 & 4.81$\pm$0.30 & 1.85$\pm$0.05 \\
    FNO-TF &  24 & 2,460,041 & 33.82$\pm$2.86 & 3.52$\pm$0.07 & 1.34$\pm$0.06 \\
    FFNO-TF & 16 & 2,855,305 & 14.13$\pm$0.59 & 2.80$\pm$0.04 & 1.15$\pm$0.01 \\
    FFNO-TF & 20 & 3,379,593 & 13.64$\pm$0.45 & 2.37$\pm$0.09 & 0.92$\pm$0.01 \\
    FFNO-TF & 24 & 3,903,881 & 12.18$\pm$0.18 & 1.98$\pm$0.02 & 0.69$\pm$0.01 \\
    \midrule
    \multicolumn{6}{l}{Our Models with FiLM only} \\
    \cmidrule(lr){2-6}
    FNO-Flow-map & 33 & 2,264,449 & 4.11$\pm$0.08 & 1.12$\pm$0.02 & 0.54$\pm$0.01\\
    \midrule
    \multicolumn{6}{l}{Our Models with Frequency-Decomposed Spectral Refinement \& FiLM} \\
    \cmidrule(lr){2-6}
    FNO-Flow-map & 24 & 1,874,313 &\textbf{2.95$\pm$0.10} & \textbf{0.89$\pm$0.04} & \textbf{0.45$\pm$0.00}\\
    F\textsuperscript{3}NO (ours) & 24 & 1,874,313 & \textbf{2.95$\pm$0.10} & \textbf{0.89$\pm$0.04} & \textbf{0.45$\pm$0.00} \\
    \bottomrule
  \end{tabular}
\vspace{-1.2em}
\end{table}

\begin{table}[b]
\vspace{-1.2em}
  \caption{[KSE] Relative $L^2$ error (\%) with mean $\pm$ standard deviation for varying $\nu$ with Frequency-Decomposed Spectral Refinement and a fixed resolution $S=32$.}
  \label{tab:kse_freq_ablation32}
  \centering
      \scriptsize
    \setlength{\tabcolsep}{6pt}
  \begin{tabular}{lccccc}
    \toprule
    \multirow{2}{*}{Model} & \multirow{2}{*}{Modes} & \multirow{2}{*}{Number of Parameters} & \multicolumn{3}{c}{$\nu$}                 \\
    \cmidrule(r){4-6}
        &  &  & 0.075 & 0.1 & 0.125 \\
    \midrule
    \multicolumn{6}{l}{Rollout} \\
    \cmidrule(lr){2-6}
    FNO-TF &  17 & 1,213,825 & 101.68$\pm$3.85 & 49.83$\pm$0.59 & 29.43$\pm$0.39 \\
    FFNO-TF & 17 & 2,788,737 & 72.93$\pm$1.14 & 26.11$\pm$0.77 & 13.58$\pm$0.29 \\
    \midrule
    \multicolumn{6}{l}{Rollout with Frequency-Decomposed Spectral Refinement} \\
    \cmidrule(lr){2-6}
    FNO-TF &  8 & 821,641 & 84.36$\pm$0.62 & 34.58$\pm$0.37 & 18.65$\pm$1.34 \\
    FNO-TF &  10 & 952,713 & 86.72$\pm$1.07 & 36.87$\pm$0.30 & 20.01$\pm$1.87 \\
    FNO-TF &  12 & 1,345,929 & 91.46$\pm$1.49 & 39.69$\pm$0.07 & 22.21$\pm$0.80 \\
    FFNO-TF & 8 & 1,806,729 & 63.20$\pm$0.90 & 19.91$\pm$0.04 & 10.11$\pm$0.28 \\
    FFNO-TF & 10 & 2,068,873 & 62.49$\pm$1.02 & 18.83$\pm$0.09 & 9.18$\pm$0.24 \\
    FFNO-TF & 12 & 2,855,305 & 63.13$\pm$1.39 & 19.21$\pm$0.02 & 8.73$\pm$1.60 \\
    \midrule
    \multicolumn{6}{l}{Our Models with FiLM only} \\
    \cmidrule(lr){2-6}
    FNO-Flow-map & 17 & 1,215,873 & 5.62$\pm$0.08 & 1.69$\pm$0.05 & 0.86$\pm$0.02\\
    \midrule
    \multicolumn{6}{l}{Our Models with Frequency-Decomposed Spectral Refinement \& FiLM} \\
    \cmidrule(lr){2-6}
    FNO-Flow-map & 12 & 1,087,881 & \textbf{3.91$\pm$0.06} & \textbf{1.26$\pm$0.03} & \textbf{0.70$\pm$0.03} \\
    F\textsuperscript{3}NO (ours) & 12 & 1,087,881 & \textbf{3.91$\pm$0.06} & \textbf{1.26$\pm$0.03} & \textbf{0.70$\pm$0.03} \\
    \bottomrule
  \end{tabular}
\end{table}

\paragraph{Frequency-decomposed spectral refinement used in autoregressive rollout.} The first two row blocks, {Rollout} and {Rollout with Frequency-Decomposed Spectral Refinement (FDSR)}, compare standard FNO/FFNO models trained with teacher forcing (TF) against their FDSR-equipped counterparts, also trained with TF. When using FDSR with 40 or 48 retained modes, where the retained modes model the low-frequency component and the remaining modes are used for high-frequency refinement, the proposed method consistently improves prediction accuracy over the standard counterparts. These results suggest that inter-frequency mixing through frequency-decomposed refinement provides clear benefits even in the standard rollout setting.


\paragraph{Ablation on the architectural choices: FiLM only or FDSR + FiLM.} To test the effectiveness of the FDSR, we compare the performance of the models with FiLM only and with FDSR + FiLM. Regardless of the choice of the backbone network and also regardless of the operation modes (i.e., flow-map or hybrid), adding FDSR provides improvement in terms of prediction accuracy as well as decreased model parameter counts.  This result suggests that the FDSR not only serves as an effective tool for mixing inter-frequency information, but also as an efficient alternative for modeling high-frequency modes. 

\paragraph{Ablation on the operation modes: Rollout or Flow-map or Hybrid.} The results clearly show that relying on a single operation mode (either Rollout or Flow-map) does not provide the best performance results. Regardless of the time-conditioning mechanisms (FiLM or FDSR+FiLM), the hybrid method (i.e., the segmented rollout) produces prediction results that outperform both baselines (rollout and flow-map). 


Tables~\ref{tab:kse_freq_ablation64}--\ref{tab:kse_freq_ablation32} report the same comparison at lower spatial resolutions, $S=64$ and $S=32$. A different pattern emerges from the high-resolution setting. FDSR continues to improve the direct flow-map predictor at both resolutions. For example, at $S=64$, adding FDSR reduces the FiLM-only flow-map error from $4.11\%$ to $2.95\%$ at $\nu=0.075$, and at $S=32$ from $5.62\%$ to $3.91\%$. Similar improvements are observed for the other viscosities. This indicates that the frequency-decomposed refinement remains useful even when the spatial representation is coarser.

In contrast, segmented execution provides little additional benefit at these lower resolutions. The direct FNO-Flow-map+FDSR and F\textsuperscript{3}NO results are nearly identical at both $S=64$ and $S=32$. This suggests that, after spatial coarsening removes much of the fine-scale complexity, a single finite-time flow-map can already model the full prediction horizon effectively. In these settings, further segmentation is unnecessary because the long-range mapping is sufficiently tractable. The benefit of Hybrid-TF therefore becomes more pronounced as spatial resolution increases, whereas at lower resolutions direct full-horizon prediction can be sufficient.

Overall, these results suggest that explicitly modeling the distinct behaviors of low- and high-frequency components provides a useful inductive bias for time-conditioned prediction.

For the FDSR variants, we set the spectral cutoff to approximately three quarters of the full modes: specifically, $m=48$ for $S=128$, $m=24$ for $S=64$, and $m=12$ for $S=32$. Empirically, this choice provides a favorable accuracy--parameter trade-off while preserving a nontrivial set of high-frequency components for the refinement branch.

\subsection{Layer-wise Contribution of High-Frequency Branch}
\label{app:fdsr_branch_ablation}

We remove the high-frequency branch of a trained F$^3$NO at inference by setting $g^{(\ell)}_{\text{high}}\!:=\!0$ in Eq.~(\ref{eq:fdsr_fusion}) for one layer at a time, or for all layers, and re-evaluate the segmented rollout at $\tau=4$. No weights change and the intact model reproduces its reported error exactly. Table~\ref{tab:fdsr_branch_ablation} gives the rollout error relative to the intact model.

Every layer's branch contributes, as removing it from any single layer raises the rollout error by $3.4$ to $23.5\times$ on KSE and $4.5$ to $13.0\times$ on NSE. Replacing the branch output by its test-set mean is as damaging as removing it, so the branch supplies input-dependent corrections. The most important layer differs by system, the middle layers on KSE and the deeper layers on NSE.

These factors measure reliance within a model trained with FDSR, not the gain from adding FDSR, which Tables~\ref{tab:kse_lowfreq_ablation128} and~\ref{tab:nse_lowfreq_ablation128} report.

\begin{table}[b]
\caption{Rollout error growth when the high-frequency branch is removed at inference ($\tau=4$; KSE $\nu=0.075$, $S=128$; NSE $\nu=10^{-5}$, $S=64\times64$). Factors are relative to F$^3$NO.}
\label{tab:fdsr_branch_ablation}
\centering
\small
\begin{tabular}{l S[table-format=2.1] S[table-format=2.1]}
\toprule
Intervention & {KSE} & {NSE} \\
\midrule
F$^3$NO & 1.0$\times$ & 1.0$\times$ \\
no high branch, layer 0 & 15.3$\times$ & 6.6$\times$ \\
no high branch, layer 1 & 10.4$\times$ & 4.5$\times$ \\
no high branch, layer 2 & 23.5$\times$ & 13.0$\times$ \\
no high branch, layer 3 & 3.4$\times$ & 9.7$\times$ \\
no high branch, all layers & 31.5$\times$ & 17.0$\times$ \\
mean branch output, all layers & 42.0$\times$ & 19.3$\times$ \\
\bottomrule
\end{tabular}
\end{table}
\section{Segmented Training and Prediction}\label{app:seg}

We examine how segmentation affects prediction accuracy across time-conditioned operators, spatial resolutions, and dynamical regimes. The segment length $\tau$ specifies the number of snapshot intervals per segment, with $\tau=1$ corresponding to one-step autoregression and a segment spanning the full horizon corresponding to direct prediction. Shorter segments limit the interval covered by each prediction but require more recursive transitions, while longer segments reduce these transitions at the cost of potentially harder finite-time mappings. We first compare time-conditioned operators under direct and segmented prediction, then examine the effects of segment length and training strategy across PDE benchmarks. Hybrid-TF uses ground-truth segment-boundary states during training, whereas Hybrid-RO propagates predicted states between training segments. Both use predicted boundary states at inference. Finally, we evaluate a complementary error-propagation diagnostic that assesses candidate execution intervals using a fixed trained model and compare its preferred interval with the empirical sweep results.

\subsection{
Effect of Segmentation across Time-Conditioned Operators
}
TCIN~\cite{herde2024poseidon} combined with FNO and FFNO (denoted as FNO-TCIN and FFNO-TCIN, respectively) provides the closest architectural baseline to our proposed method, as TCIN also employs hypernetwork-based FiLM-style shift-and-scale modulation for time-displacement conditioning. 
To examine whether the benefit of segmentation is specific to F\textsuperscript{3}NO or extends to other time-conditioned operators, we apply the same Hybrid-TF strategy to FNO-TCIN, FFNO-TCIN, an FNO with FiLM conditioning, and F\textsuperscript{3}NO. Table~\ref{tab:tcin-f3no} compares each model with its corresponding direct flow-map counterpart.

\begin{table}[t]
  \caption{Relative $L^2$ error (\%) with mean $\pm$ standard deviation for varying datasets and varying PDE parameters with a resolution $S=128$ for BE/KSE, a resolution $S=64\times64$ for NSE and SWE, and a resolution $S=32\times32\times32$ for RTI. The hybrid segmented execution uses $\tau=4$ for KSE, NSE, and RTI, $\tau=19$ for BE, and $\tau=10$ for SWE.}
  \label{tab:tcin-f3no}
  \centering
      \scriptsize
    \setlength{\tabcolsep}{2pt}
  \begin{tabular}{lccccccccc}
    \toprule
     & \multicolumn{1}{c}{BE} & \multicolumn{3}{c}{KSE} & \multicolumn{3}{c}{NSE} & \multicolumn{1}{c}{SWE}  &  \multicolumn{1}{c}{RTI}            \\
         & $\nu=0.001$ & $\nu=0.075$ & $\nu=0.1$ & $\nu=0.125$ & $\nu=10^{-5}$ & $\nu=10^{-4}$ & $\nu=10^{-3}$ & $r\sim\mathcal{U}(0.3,0.7)$ & $A_t=0.25$\\
    \midrule
    \multicolumn{9}{l}{Flow-map} \\
    \cmidrule(lr){2-10}
    FNO-TCIN& 
    91.68$\pm$0.62 & 8.24$\pm$0.05 & 3.38$\pm$0.03 & 2.00$\pm$0.01 & 9.72$\pm$0.52 &7.71$\pm$0.17 & 2.19$\pm$0.09 & 0.25$\pm$0.05 & 5.15$\pm$0.03\\
    FFNO-TCIN& 
    1.88$\pm$0.07 & 2.84$\pm$0.01 & 1.01$\pm$0.04 & 0.59$\pm$0.09 & 3.03$\pm$0.14 & 3.25$\pm$0.07 & 1.73$\pm$0.13 & 0.26$\pm$0.00 & 5.12$\pm$0.02 \\
    FNO with FiLM only & 2.24$\pm$0.05 & 3.38$\pm$0.06 & 0.90$\pm$0.04 & 0.43$\pm$0.00 & 5.77$\pm$0.36 & 4.90$\pm$0.09 & 1.22$\pm$0.07 & 0.26$\pm$0.01 & 5.06$\pm$0.01\\
    F\textsuperscript{3}NO (w/o hybrid) & 
    \textbf{1.37$\pm$0.11} &2.01$\pm$0.05 & 0.63$\pm$0.01 & 0.34$\pm$0.00  & 4.88$\pm$0.11 & 4.79$\pm$0.50 & 1.08$\pm$0.08 & 0.23$\pm$0.01 & 4.98$\pm$0.01\\
    \midrule
    \multicolumn{9}{l}{Flow-map+Hybrid-TF} \\
    \cmidrule(lr){2-10}
    FNO-TCIN& 
    90.94$\pm$0.78 & 11.68$\pm$0.32 & 5.16$\pm$0.67 & 2.95$\pm$0.11 & 3.87$\pm$0.09 & 2.32$\pm$0.09 & 1.12$\pm$0.03 & 0.12$\pm$0.00 & 5.44$\pm$0.13\\
    FFNO-TCIN& 
    1.78$\pm$0.05 & 1.84$\pm$0.05 & 0.81$\pm$0.03 & 0.41$\pm$0.08 & 1.94$\pm$0.04 & 2.04$\pm$0.15 & 0.72$\pm$0.02 & 0.19$\pm$0.01 & 4.97$\pm$0.10\\
    FNO with FiLM only & 2.30$\pm$0.04 & 1.40$\pm$0.05 & 0.44$\pm$0.01 & 0.22$\pm$0.01 & 1.68$\pm$0.00 &0.95$\pm$0.04 & 0.24$\pm$0.00 & 0.09$\pm$0.01 & 4.93$\pm$0.04\\
    F\textsuperscript{3}NO & 
    \textbf{1.37$\pm$0.11} & \textbf{1.15$\pm$0.06} & \textbf{0.32$\pm$0.00} & \textbf{0.17$\pm$0.01} & \textbf{1.44$\pm$0.04} & \textbf{0.58$\pm$0.02} & \textbf{0.22$\pm$0.00} & \textbf{0.07$\pm$0.01} & \textbf{4.83$\pm$0.08}\\
    \bottomrule
  \end{tabular}
\end{table}

\begin{figure}[b]
    \vspace{-1.2em}
        \centering
        \includegraphics[width=0.8\linewidth]{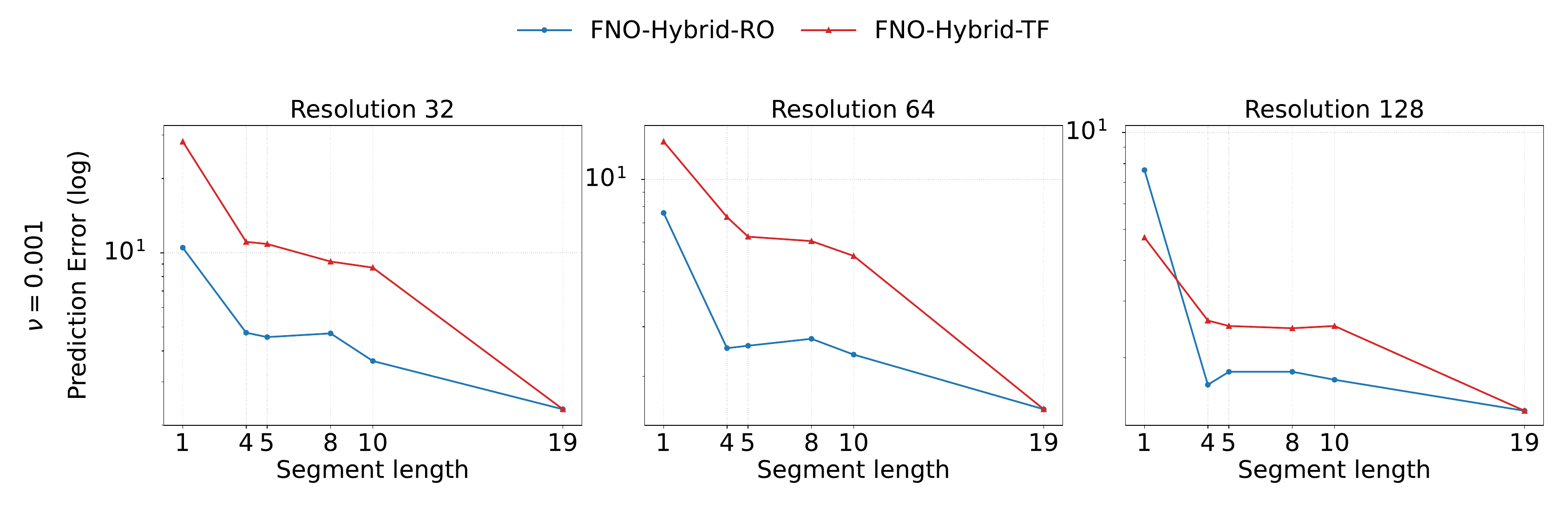}
    \vspace{-0.6em}    
    \caption{[BE] Prediction error for the proposed hybrid model as a function of segmentation length with $\nu = 0.001$ across three spatial resolution, $S=32$, $S=64$, and $S=128$.}
    \label{fig:burgers_loss_seg}
\end{figure}

The effect of segmentation depends on the underlying time-conditioned operator. For FFNO-TCIN, Hybrid-TF reduces the prediction error for all benchmark problems. The FiLM-only FNO and F\textsuperscript{3}NO show a similar pattern, with particularly large reductions on the longer and more challenging dataset. For F\textsuperscript{3}NO, for example, the mean error decreases from $2.01\%$ to $1.15\%$ on KSE with $\nu=0.075$, and from $4.88\%$ to $1.44\%$ on NSE with $\nu=10^{-5}$. For BE, we use $\tau=19$, which covers the full prediction horizon. The corresponding Hybrid-TF configuration contains only a single segment and is equivalent in execution to direct flow-map prediction.

These results show that segmentation can improve time-conditioned finite-time prediction when the full-horizon mapping is difficult. Under the segmented settings, F\textsuperscript{3}NO achieves the lowest prediction error in every reported multi-segment configuration. The comparison with the direct prediction suggests that the gains from FDSR and Hybrid-TF are complementary.

\subsection{
Effect of Segment Length and Training Strategy
}

\begin{figure}[t]
        \centering
        \includegraphics[width=0.8\linewidth]{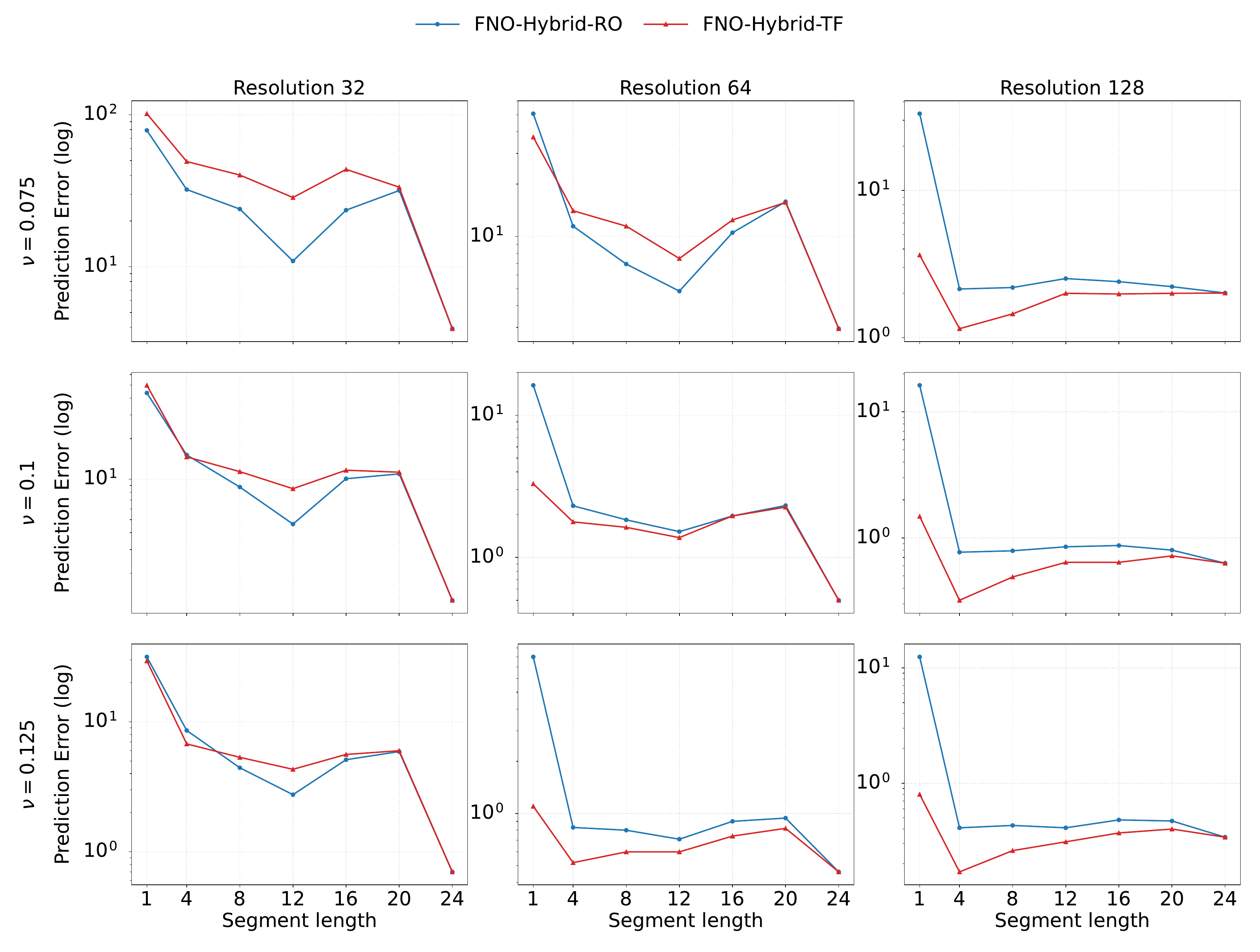}
    \vspace{-0.6em}
    \caption{[KSE] Prediction error for the proposed hybrid model as a function of segmentation length with $\nu = 0.075$, $\nu = 0.1$, and $\nu = 0.125$ across three spatial resolution, $S=32$, $S=64$, and $S=128$.}
    \label{fig:KS_loss_seg}
    \vspace{-1.2em}
\end{figure}

\begin{figure}[b]
\vspace{-1.2em}
        \centering
        \includegraphics[width=0.8\linewidth]{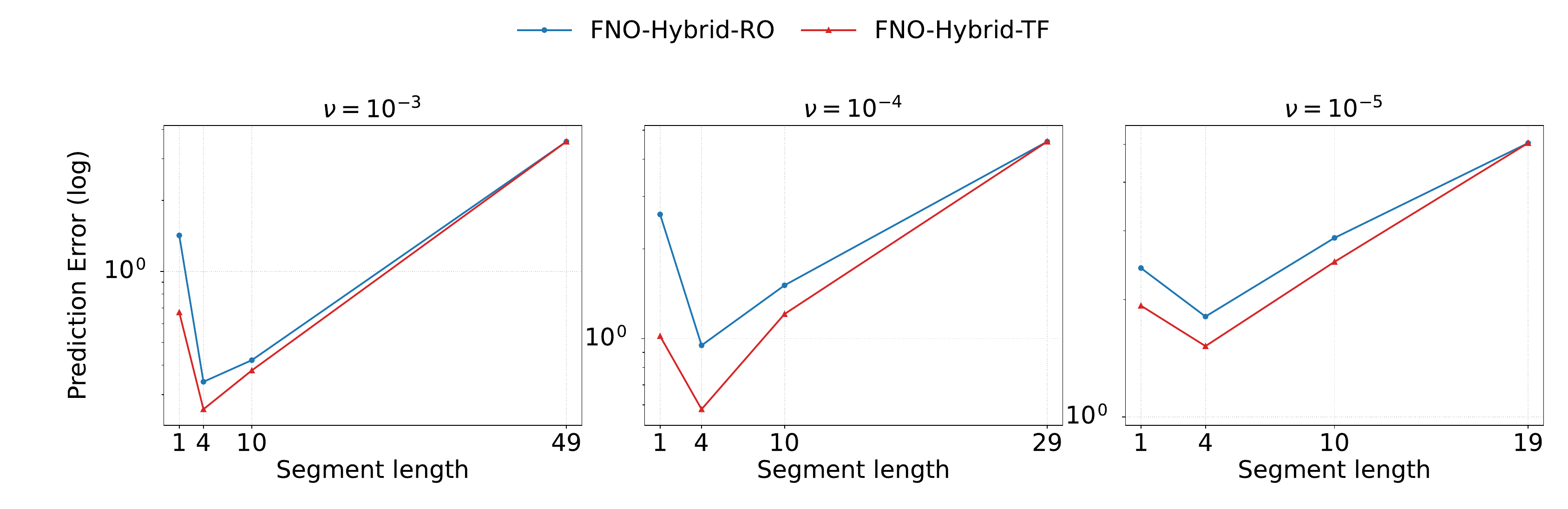}
    \vspace{-0.6em}
    \caption{[NSE] Prediction error for the proposed hybrid model as a function of segmentation length with $\nu = 10^{-3}$, $\nu = 10^{-4}$, and $\nu = 10^{-5}$ with a fixed resolution $S=64\times64$.}
    \label{fig:NS_loss_seg}
\end{figure}

\begin{figure}[!t]
    \centering
    \hspace*{\stretch{1}}%
    \begin{subfigure}[t]{0.48\linewidth}
        \centering
        \includegraphics[width=0.57\linewidth]{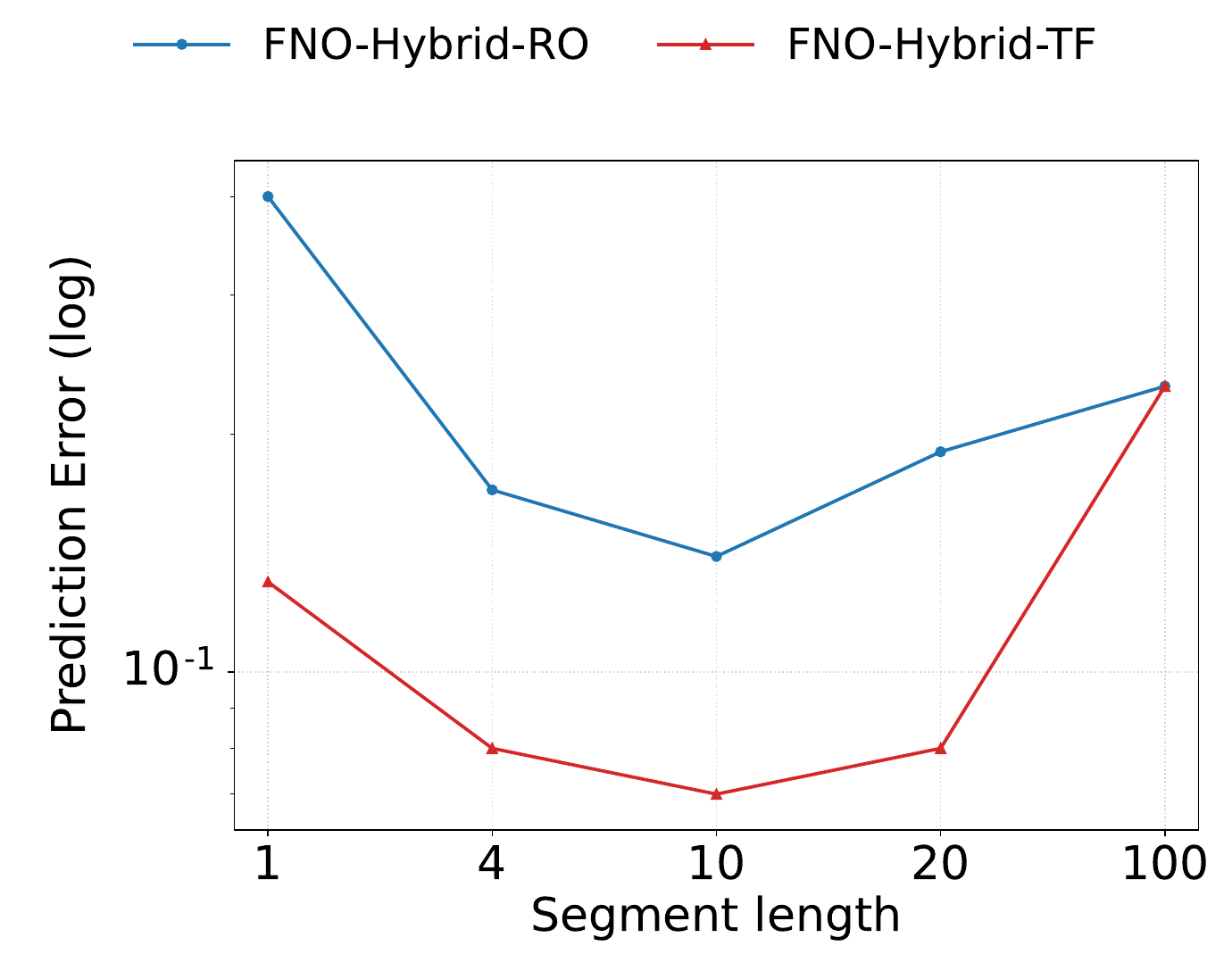}
        \caption{[SWE] Prediction error for the proposed hybrid model as a function of segmentation length with a fixed resolution $S=64\times64$.}
        \label{fig:SWE_loss_seg}
    \end{subfigure}%
    \hspace{\stretch{2}}%
    \begin{subfigure}[t]{0.48\linewidth}
        \centering
        \includegraphics[width=0.57\linewidth]{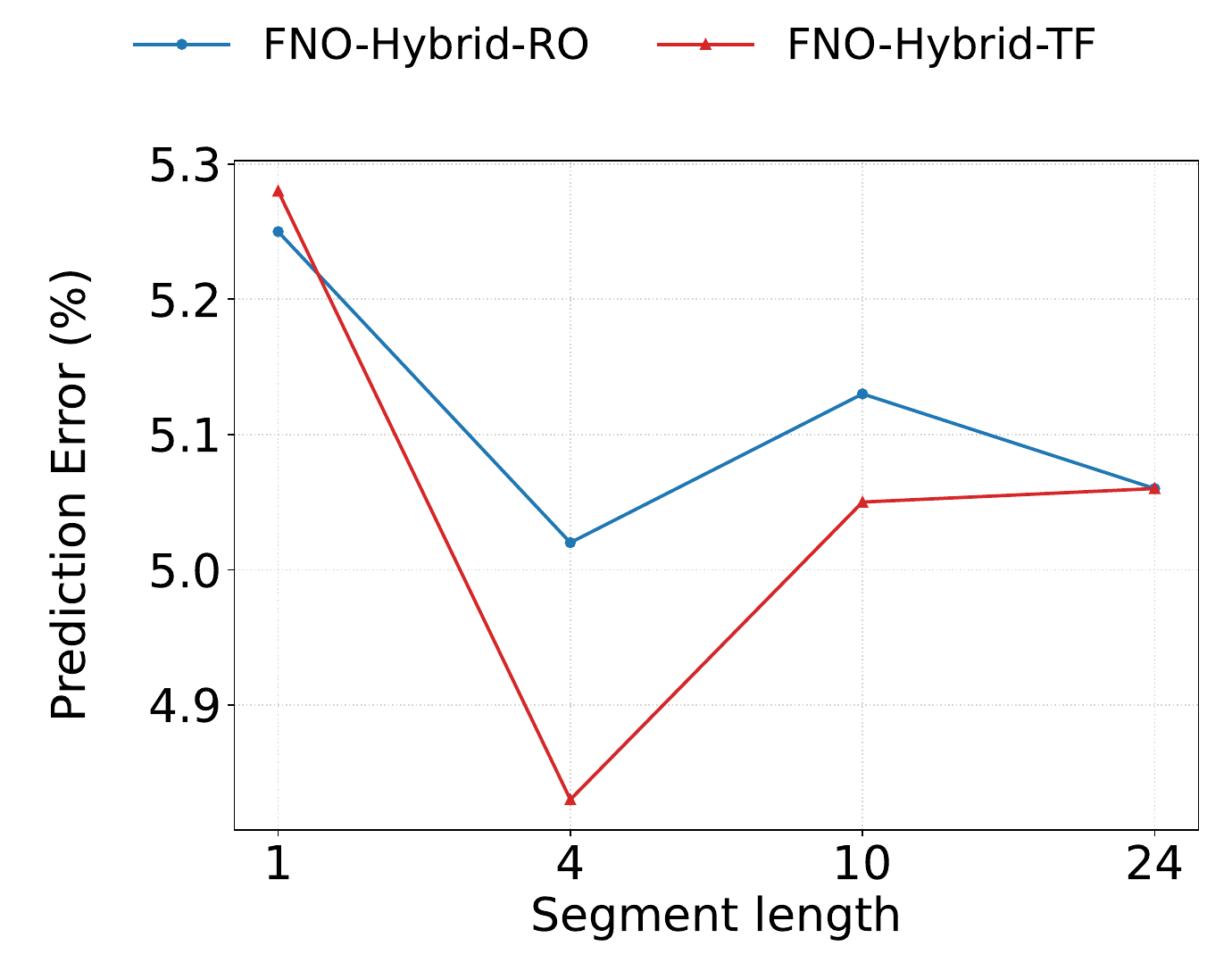}
        \caption{[RTI] Prediction error for the proposed hybrid model as a function of segmentation length with a fixed resolution $S=32\times32\times32$}
        \label{fig:RTI_loss_seg}
    \end{subfigure}%
    \hspace*{\stretch{1}}
    \caption{[SWE/RTI] Prediction error for the proposed hybrid model as a function of segmentation length.}
    \label{fig:swe_rti_loss_seg}
\end{figure}

We further evaluate the proposed segmented execution strategy on BE, KSE, NSE, SWE, and RTI benchmarks, as shown in Figures~\ref{fig:burgers_loss_seg}--\ref{fig:swe_rti_loss_seg}. All considered methods are variants of F\textsuperscript{3}NOs and the naming convention is 
FNO-Hybrid-[operation-mode], where the
operation-mode $\in$ \{RO, TF\}. 
In all five benchmarks, the case $\tau=1$ corresponds to the pure autoregressive rollout, while the largest segment length in each plot, i.e., $\tau=19$ for BE, $\tau=24$ for KSE and RTI, $\tau=19, 29, 49$ for the three NSE viscosity ($10^{-5},10^{-4},10^{-3}$), and $\tau=100$ for SWE, corresponds to the single global finite-time flow-map prediction over the full available horizon. These results highlight how the optimal segment length depends on both the underlying dynamics and the prediction horizon. Across the BE and KSE benchmarks, the pure rollout generally produces the largest prediction errors, confirming that stepwise autoregression is vulnerable to long-horizon error accumulation. For these 1D systems, increasing the segment length often improves performance, and the global flow-map can be competitive, especially at lower spatial resolutions and at 
higher viscosity, where the dynamics are smoother and the long-range mapping is easier to approximate. 

In the NSE benchmark with chosen specific viscosity values ($10^{-5},10^{-4},10^{-3}$), the resulting dynamics appear to be more complex and both extremes (i.e., pure rollout and pure flow-map) become less optimal: $\tau=1$ suffers from recursive error accumulation, while the full-horizon segment also yields large errors because the model must directly learn a highly nonlinear mapping between temporally distant and weakly correlated states. Intermediate segment lengths therefore achieve better performance by balancing these two sources of error. Similar results are observed in the SWE and RTI benchmarks. 


\textbf{Segmented Autoregressive Rollout or Segmented Teacher Forcing}:
The results also compare two segmented execution strategies: Hybrid-RO and Hybrid-TF. Hybrid-RO trains the model under a segmented rollout setting, where predicted segment-boundary states are recursively used to initialize subsequent segments. This setting better matches long-horizon inference, but it is more sensitive to boundary prediction errors. Hybrid-TF, in contrast, re-grounds the model with ground-truth segment-boundary states during training, which can stabilize optimization by preventing early errors from contaminating later segments. The comparisons between Hybrid-RO and Hybrid-TF further show that the preferred training strategy depends on the complexity of the target dynamics. Hybrid-RO is mainly competitive in relatively easier settings, such as BE and the low-resolution KSE case, where predicted segment-boundary states can be propagated with small drift. In more challenging settings, including higher resolution KSE and the two-dimensional NSE benchmark, Hybrid-TF often yields lower or more stable errors by re-grounding segment boundaries during training, which helps prevent early segment-boundary errors from accumulating across subsequent segments. SWE and RTI benchmarks show a similar trend to the NSE benchmark, with Hybrid-TF being the robust choice in the more challenging long-horizon settings.

Overall, these results support the motivation for segmented hybrid execution. Rather than committing to either pure rollouts or a single global flow-map, the proposed strategy provides a trade-off between the two. This allows the model to reduce stepwise error accumulation while avoiding the optimization difficulty of overly long one-shot finite-time prediction.

\subsection{Error Propagation in Autoregressive and Segmented Prediction}

\begin{table}[t]
\caption{[KSE, $\nu=0.075$, $S=128$] Direct error per snapshot interval, $\eta(j)$, normalized by the minimum error within each trained-\(\tau\) model. For all models, the minimum error occurs at $j=4$, consistent with the empirical choice to run KSE $S=128$ at $\tau=4$.}
\label{tab:kse_error_propagation}
\vspace{-0.6em}
\centering
\small
\begin{tabular}{lccccccccc}
\toprule
Trained $\tau$ & $j{=}1$ & 2 & 4  & 6 & 8  & 10 & 12 & 16 & 24 \\
\midrule
4 & 1.60 & 1.18 & \textbf{1.00} & -- & -- & -- & -- & -- & --  \\
8 & 2.35 & 1.31 & \textbf{1.00}  & 1.25 & 2.10 & -- & -- & -- & -- \\
10 & 2.40& 1.32 & \textbf{1.00} & 1.02 & 1.15 & 2.43 & -- & -- & --\\
12 & 2.47 & 1.48 & \textbf{1.00}  & 1.03 & 1.27 & 2.58 & 3.22 & -- & -- \\
24 & 2.70 & 1.62 & \textbf{1.00}  & 1.02 & 1.26 & 1.67 & 2.17 & 2.75 & 2.62 \\
\bottomrule
\end{tabular}
\end{table}

\begin{table}[b]
\caption{[BE, $\nu=0.001$, $S=128$] Direct error per snapshot interval, $\eta(j)$, normalized by the minimum error within each trained-\(\tau\) model. 
For each trained-\(\tau\) model, the minimum of $\eta(j)$ occurs at its longest supported lead; the full-horizon model attains its minimum at j=19, consistent with the empirical choice $\tau=19$.
}
\vspace{-0.6em}
\label{tab:be_error_propagation}
    \centering
    \small
    \begin{tabular}{lccccccccc}
        \toprule
        Trained $\tau$ & $j{=}1$ & 2 & 4 & 6 & 8 & 10 & 12 & 16 & 19\\
        \midrule
        2  & 1.30 & \textbf{1.00} & -- & -- & -- & -- & -- & -- & -- \\
        4  & 1.76 & 1.19 & \textbf{1.00} & -- & -- & -- & -- & -- & -- \\
        8  & 2.86 & 1.95 & 1.32 & 1.24 & \textbf{1.00} & -- & -- & -- & -- \\
        10 & 3.78 & 2.87 & 1.73 & 1.47 & 1.12 & \textbf{1.00} & -- & -- & -- \\
        12 & 4.55 & 2.86 & 1.86 & 1.69 & 1.42 & 1.23 & \textbf{1.00} & -- & -- \\
        16 & 5.67 & 3.62 & 2.81 & 2.96 & 3.28 & 2.99 & 2.25 & \textbf{1.00} & -- \\
        19 & 9.16 & 8.14 & 7.50 & 7.50 & 6.99 & 6.12 & 5.19 & 3.00 & \textbf{1.00} \\
        \bottomrule
    \end{tabular}
\end{table}

\label{sec:error_propagation}
The main results in Table~\ref{tab:varying_data_S128} use segment lengths selected through empirical sweeps. Here, we provide a complementary analysis of 
whether the same preferred execution interval can be identified from the direct finite-time prediction errors of a trained model.
Since F\textsuperscript{3}NO supports predictions at different lead times, this criterion can be evaluated from a single trained model without retraining for each candidate segment length. 
We derive a simple error-propagation criterion that evaluates candidate intervals from ground-truth segment-start states. This avoids running a full segmented rollout for every candidate interval. Across all benchmark problems, the interval preferred by this criterion agrees with the empirical segment-length sweep.

The segment length $\tau$ sets how often predicted states are fed back and how far each prediction must reach. Short segments make each prediction easier but require more recursive transitions, and every transition contributes its own error to the rollout \cite{brandstetter2022message,stachenfeld2021learned,liu2022hierarchical,mccabe2023towards}. Longer segments reduce the number of recursive transitions, but require the model to predict across larger temporal gaps, which can increase the direct prediction error \cite{liu2022hierarchical,lippe2023pde,bi2023accurate}. Consistent with this trade-off, intermediate step sizes have been found to give the lowest rollout error in learned simulators and forecasting models \cite{stachenfeld2021learned,li2022learning,andrae2025continuous}, and the same balance arises in classical numerical integration between truncation error and accumulated round-off error \cite{atkinson2009numerical}. 
We make this trade-off explicit with a simple error-propagation bound and use it to identify the quantity that determines the preferred segment length, and report that quantity for all benchmark problems.

\begin{table}[t]
\caption{[NSE, $\nu=10^{-5}$, $S=64\times64$] Direct error per snapshot interval, $\eta(j)$, normalized by the minimum error within each trained-\(\tau\) model. For all models, the minimum error occurs at $j=4$, consistent with the empirical choice to run NSE $S=64\times64$ at $\tau=4$.}
\label{tab:nse_error_propagation}
\vspace{-0.6em}
    \centering
    \small
    \begin{tabular}{lccccccccc}
        \toprule
        Trained $\tau$ & $j{=}1$ & 2 & 4 & 6 & 8 & 10 & 12 & 16 & 19\\
        \midrule
        4 & 1.45 & 1.07 & \textbf{1.00} & -- & -- & -- & -- & -- & --\\
        8 & 1.63 & 1.21  & \textbf{1.00} & 1.04 & 1.20 & -- & -- & -- & --\\
        12 & 1.64 & 1.27  & \textbf{1.00} & 1.03 & 1.03 & 1.10 & 1.20 & -- & --\\
        16 & 1.64 & 1.20  & \textbf{1.00} & 1.02 & 1.10 & 1.21 & 1.33 & 1.68 & --\\
        19 & 1.59 & 1.24  & \textbf{1.00} & 1.01 & 1.06 & 1.15 & 1.24 & 1.46 & 1.58\\
        \bottomrule
    \end{tabular}
\end{table}

\begin{table}[b]
\caption{[SWE, $r\sim\mathcal{U}(0.3,0.7)$, $S=64\times64$] Direct error per snapshot interval, $\eta(j)$, normalized by the minimum error within each trained-\(\tau\) model. For all models, the minimum error occurs at $j=10$, consistent with the empirical choice to run SWE $S=64\times64$ at $\tau=10$.}
\label{tab:swe_error_propagation}
\vspace{-0.6em}
    \centering
    \small
    \begin{tabular}{lcccccccc}
        \toprule
        Trained $\tau$ & $j{=}1$ & 2 & 4 & 5 & 10 & 20 & 25 & 50\\
        \midrule
        10 & 1.90 & 1.47 & 1.18 & 1.11 & \textbf{1.00} & -- & -- & -- \\
        20 & 4.46 & 1.36 & 1.09 & 1.05 & \textbf{1.00} & 1.08 & -- & -- \\
        50 & 2.55 & 1.56 & 1.07 & 1.03 & \textbf{1.00} & 1.02 & 1.03 & 1.06 \\
        \bottomrule
    \end{tabular}
\end{table}

\begin{table}[t]
\caption{[RTI, $A_{t}=0.25$, $S=32\times32\times32$] Direct error per snapshot interval, $\eta(j)$, normalized by the minimum error within each trained-\(\tau\) model. For all models, the minimum error occurs at $j=4$, consistent with the empirical choice to run RTI $S=32\times32\times32$ at $\tau=4$.}
\label{tab:rti_error_propagation}
\vspace{-0.6em}
    \centering
    \small
    \begin{tabular}{lccccccccc}
        \toprule
        Trained $\tau$ & $j{=}1$ & 2 & 4 & 6 & 8 & 10 & 12 & 16 & 24\\
        \midrule
        4  & 1.42 & 1.04 & \textbf{1.00} & -- & -- & -- & -- & -- & -- \\
        10 & 1.37 & 1.04 & \textbf{1.00} & 1.08 & 1.36 & 2.11 & -- & -- & -- \\
        16 & 1.99 & 1.12 & \textbf{1.00} & 1.14 & 1.39 & 1.98 & 2.78 & 4.13 & -- \\
        24 & 2.40 & 1.22 & \textbf{1.00} & 1.48 & 2.07 & 3.41 & 4.97 & 7.25 & 7.39 \\
        \bottomrule
    \end{tabular}
\end{table}
Let $h_{\mathcal D}$ be the snapshot interval and $T=Nh_{\mathcal D}$ the prediction horizon. As in Section~\ref{sec:hybrid}, Let $\tau\ge1$ denote the number of snapshot intervals per segment, corresponding to a duration $\tau h_{\mathcal D}$. We partition the prediction interval $[0,T]$ into
$K=\lceil T/(\tau h_{\mathcal D})\rceil$ segments with boundaries
\[
0=t_0<t_1<\cdots<t_K=T,
\qquad
t_{k+1}-t_k\le\tau h_{\mathcal D}.
\]
Within a segment, every state is predicted directly from the segment's start state, and only the terminal prediction is passed to the next segment. Let
\[
\delta_k=t_{k+1}-t_k\le \tau h_{\mathcal D}
\]
denote the duration of segment $k$. Let $\Phi_{\delta}$ denote the exact solution operator over a time interval $\delta$, and let $F_{\delta}(u):=\widehat F_\theta(u,\delta)$ denote the corresponding learned finite-time predictor. We define the error at the $k$-th segment boundary as
\[
E_k=\|\widehat{u}_{t_k}-u_{t_k}\|,
\qquad E_0=0.
\]
Assume that $F_{\delta_k}$ is $L_{\delta_k}$-Lipschitz on the relevant states and that its direct error from a true state is bounded,
\begin{equation}
    \|F_{\delta_k}(u_{t_k})-\Phi_{\delta_k}(u_{t_k})\| \le \epsilon_{\delta_k} .
\end{equation}
Adding and subtracting $F_{\delta_k}(u_{t_k})$ gives
\begin{equation}
    E_{k+1} \le \|F_{\delta_k}(\widehat{u}_{t_k})-F_{\delta_k}(u_{t_k})\| + \|F_{\delta_k}(u_{t_k})-\Phi_{\delta_k}(u_{t_k})\| \le L_{\delta_k}E_k+\epsilon_{\delta_k},
\end{equation}
and therefore
\begin{equation}
    E_K \le \epsilon_{\delta_j} \sum_{j=0}^{K-1} \prod_{\ell=j+1}^{K-1}L_{\delta_\ell} .
\end{equation}
If the predictor is non-expansive, $L_{\tau h_{\mathcal D}}\le 1$, the bound reduces to $E_K \le K\epsilon_{\tau h_{\mathcal D}}$. The bound is then smallest for the segment length that minimizes the direct error per snapshot interval, $\eta(\tau) = \frac{\epsilon_{\tau h_{\mathcal D}}}{\tau}$.

We evaluate $\eta$ on KSE ($\nu = 0.075$, $S = 128$) over a horizon of 24 snapshot intervals. F$^3$NO is trained with $\tau \in \{4, 8, 10, 12, 24\}$ and otherwise the configuration of Section~\ref{sec:exp_setup}, with three seeds per setting. For each lead $j$ up to the training segment length, $\epsilon_j$ is measured as the relative $L^2$ error of the direct prediction from the ground-truth segment-start states, averaged over the test trajectories, segment starts and seeds. Table~\ref{tab:kse_error_propagation} reports $\eta(j) = \epsilon_j/j$ relative to each model's minimum\footnote{In the experiments, $\epsilon_j$ denotes the empirical mean relative $L^2$ direct-prediction error, rather than the absolute-error upper bound used in the theoretical analysis. We use this empirical quantity as a proxy for the theoretical approximation-error term; $\eta(j)=\epsilon_j/j$ therefore serves as a heuristic criterion for comparing execution intervals.}.

In every case $\eta$ is smallest at 
$j=4$ across all tested training segment lengths,
in agreement with the segment length $\tau = 4$ used for KSE at $S = 128$ and matching the empirical sweep in Figure~\ref{fig:prediction_errorVSseg_length}. Since each curve is obtained from a single trained model, the direct error per snapshot interval provides a way to choose the segment length at this resolution without training a model for every candidate $\tau$. We perform the same analysis on the other four benchmarks and observe the same pattern, where the preferred interval identified by $\eta(j)$ agrees with the empirical segment-length sweep. The minimizer is stable across the tested training segment lengths for KSE, NSE, SWE, and RTI, while BE consistently favors longer leads and ultimately the full-horizon interval. These results suggest that the direct error per snapshot interval can help identify a preferred execution interval without running a full segmented rollout for every candidate interval.

\subsection{Irregularly Sampled Measurements}
\label{sec:irregular}
We further evaluate robustness under irregular temporal sampling, a common setting in which measurements are missing due to non-uniform acquisition. While rollouts can still be applied by advancing between observed states, their reliance on stepwise continuity makes them increasingly sensitive to missing data. In contrast, flow-map-based models naturally map between states separated by arbitrary time gaps, making them better suited for irregularly sampled trajectories. This experiment assesses whether the proposed segmented flow-map retains this advantage under substantial temporal sparsity.

Starting from fully sampled KSE trajectories, we randomly remove a fixed fraction of time steps from each trajectory, with missing percentages 
20\%, 40\%, 60\% and 80\%. The remaining observations are used as inputs, and models are tasked with predicting the system evolution across the resulting non-uniform temporal gaps. 

\begin{figure}[b]
        \centering
        \includegraphics[width=0.5\linewidth]{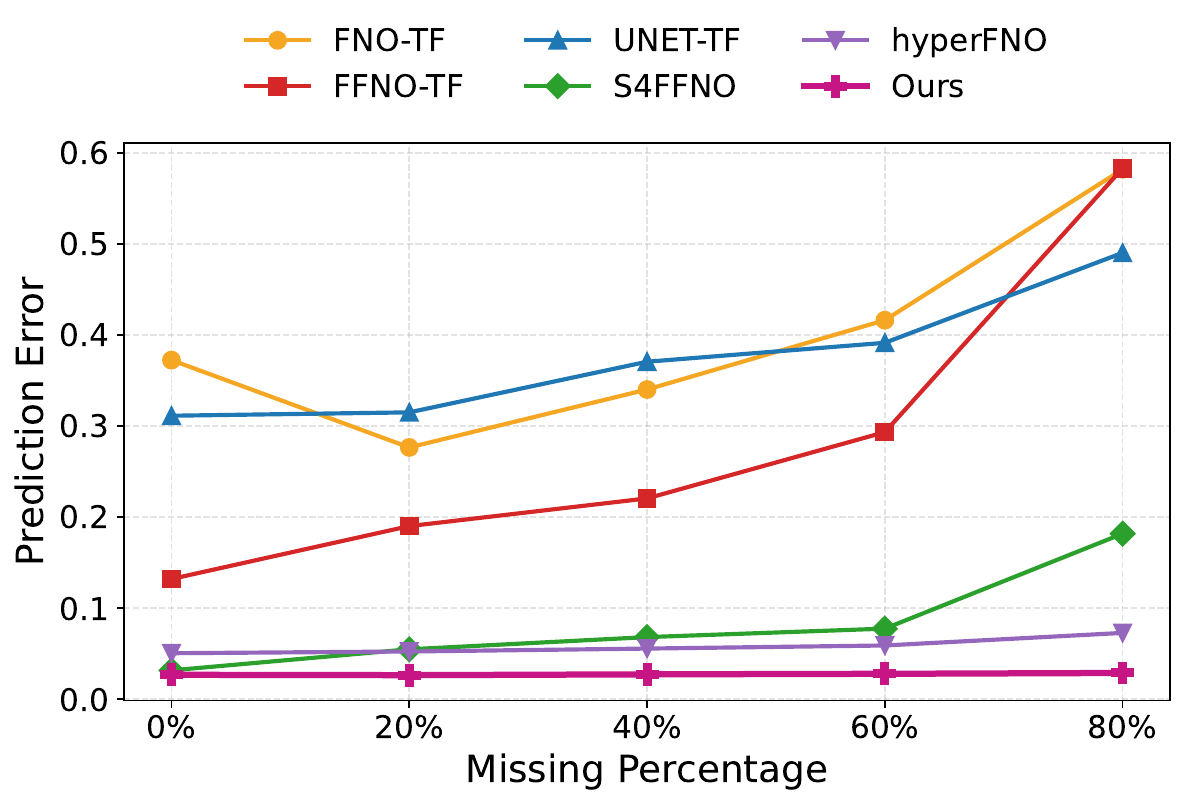}
    \caption{[KSE, $\nu=0.075$, $S=64$] Relative $L^2$ error versus missing time-step percentage under irregular temporal sampling. Missing masks are sampled independently per test trajectory, with $t=0$ and the final prediction time retained. \textbf{Ours} refers to our main FNO-Hybrid-TF model.}
    \label{fig:v0075S64_irregular}
\end{figure}

Figure~\ref{fig:v0075S64_irregular} reports the test error as a function of the missing percentage. As the fraction of missing data increases, all methods exhibit performance degradation; however, rollout-based approaches degrade rapidly due to accumulated error across increasingly long effective time steps. Direct flow-map baselines show improved robustness but still struggle when the temporal gaps become large. In contrast, the proposed segmented flow-map consistently achieves the lowest error across all missing rates and degrades most gracefully as sparsity increases. This behavior highlights a key strength of the proposed approach: by combining finite-time flow-map prediction with controlled segmentation, the method can reliably bridge irregular temporal gaps while avoiding the instability inherent to dense rollouts.

\section{Limitations}
Despite the promising empirical performance, the proposed framework has several limitations that warrant further investigation. First, the segmentation length $\tau$ in the hybrid execution strategy is currently treated as a user-defined hyperparameter, and its optimal value can vary depending on the underlying dynamics and spatial resolution. Second, the frequency decomposition relies on a fixed spectral cutoff, whereas adaptive frequency partitioning may provide additional flexibility across different PDE regimes. Finally, although the proposed framework exhibits stable and accurate long-horizon prediction empirically, further theoretical understanding of its stability and error propagation properties would be valuable.

\end{document}